\documentclass[conference]{IEEEtran}
\IEEEoverridecommandlockouts
\usepackage{amsmath,amssymb,amsfonts}
\usepackage{graphicx}
\usepackage{textcomp}
\usepackage{xcolor}
\usepackage{booktabs} % For formal tables
\usepackage[linesnumbered,ruled,vlined]{algorithm2e}
\usepackage{booktabs}
\usepackage{amsmath}
\usepackage{graphicx}
\usepackage{subfigure}
\usepackage{cite}
\usepackage{adjustbox}
\usepackage{multirow}
\usepackage{amsfonts,bm}
\usepackage{color}
\usepackage{braket}
\usepackage{mathtools}
\usepackage{enumerate}
\usepackage{enumitem}
\usepackage{empheq}
\usepackage{calc}
\usepackage{balance}
\usepackage{dsfont}
\usepackage{textcomp} 
\usepackage{color, colortbl}
\usepackage[first=0,last=9]{lcg}
\usepackage[colorlinks,linkcolor=black,citecolor=green,  urlcolor=black]{hyperref}
 
\def\BibTeX{{\rm B\kern-.05em{\sc i\kern-.025em b}\kern-.08em
    T\kern-.1667em\lower.7ex\hbox{E}\kern-.125emX}}
\begin{document}

% \title{BrainPuzzle: A New Data-Driven Method for\\ Ultrasound Brain Imaging }
\title{BrainPuzzle: Hybrid Physics and Data-Driven Reconstruction for Transcranial Ultrasound Tomography}

\author{Shengyu Chen$^{*}$, Shihang Feng$^{*}$, Yi Luo, Xiaowei Jia, and Youzuo Lin

        % <-this % stops a space
\thanks{S. Chen and S. Feng contributed equally to this work.}
\thanks{S. Chen and Y. Lin are with Los Alamos National Laboratory, Los Alamos, NM 87545, USA.}
\thanks{S. Chen and X. Jia are with University of Pittsburgh,\,\, PA, 15213, USA.}% <-this % stops a space
\thanks{Y. Luo is with Seiswave Corp,\,\, Katy, TX, 77450, USA.}% <-this % stops a space
\thanks{Y. Lin and S. Feng are with the University of North Carolina at Chapel Hill, Chapel Hill, NC 27599, USA. (E-mail: \textbf{yzlin@unc.edu}).}% <-this % stops a space
% <-this % stops a space

}

\maketitle

\begin{abstract}
Current ultrasound imaging techniques face significant challenges in producing clear brain images, primarily due to the mismatch in speed of sound between the skull and brain tissues, as well as the difficulty of effectively coupling large probes with the skull. In this paper, our goal is quantitative transcranial ultrasound: recovering a high-quality \emph{speed-of-sound}~(SoS) map of the brain. Purely physics-based full-waveform inversion (FWI) faces two practical obstacles in the cranial setting: \textbf{(i) weak signals} due to skull-induced attenuation, mode conversion, and phase aberration; and \textbf{(ii) incomplete coverage}, since full-aperture arrays are clinically and manufacturing-wise infeasible. On the other hand, fully end-to-end learning from raw channel data struggles to learn strongly nonlocal, nonlinear wave physics through bone. It often yields anatomically plausible but quantitatively biased SoS under low SNR and sparse aperture. We present \emph{BrainPuzzle}, a \textbf{hybrid} two-stage framework. \textbf{Stage~1} applies reverse time migration (TRA; time-reversal acoustics) to multi-angle acquisitions to form \emph{migration fragments} that are structurally faithful even at low SNR. \textbf{Stage~2} performs \emph{machine-learning mapping} from TRA fragments to a coherent, quantitative SoS image via a transformer-based super-resolution encoder–decoder augmented with a graph-based attention unit~(GAU) that models inter-fragment relations. A key design choice is a \textbf{partial-array} acquisition, implemented by a movable low-count transducer set, which improves feasibility and coupling while the hybrid algorithm compensates for the missing aperture. Experimental results on two distinct synthesized datasets demonstrate the superior performance of BrainPuzzle in reconstructing complete brain images from fragmented ultrasound data, highlighting its potential to advance the field of ultrasound brain image reconstruction.

\end{abstract}


\section{Introduction}

Brain imaging is central to modern healthcare, enabling non-invasive assessment of neurological disorders, monitoring disease progression, and guiding treatment decisions. Magnetic resonance imaging~(MRI) is widely regarded as the gold standard for structural and functional brain imaging due to its superior soft tissue contrast and versatility~\cite{Neuroscience-2024-Triana, Twenty-2012-Bandettini}. Computed tomography~(CT), while lower in resolution, remains the first-line modality in acute care, particularly for detecting hemorrhage and trauma, because of its speed and accessibility~\cite{Impact-2024-Choi, Computed-2022-Hillal}. Despite their clinical utility, both MRI and CT face significant limitations: MRI is costly, slow, and contraindicated for some patients, while CT exposes patients to ionizing radiation and relies on bulky, immobile equipment. These barriers can delay timely diagnosis in critical scenarios such as stroke, where rapid imaging is essential for effective intervention. This creates a pressing need for portable, real-time brain imaging modalities that can be deployed at the point of care.

One promising avenue for addressing the limitations of MRI and CT in diagnosing neurological conditions, particularly when speed, portability, and patient constraints are paramount, is ultrasound imaging~\cite{Transcranial-2013-Naqvi, wells2006ultrasound}, known for its non‐invasive nature and extensive utility in obstetrics, cardiology, and abdominal medicine, among other fields. In these areas, ultrasound has proven invaluable for monitoring fetal development, assessing cardiac function, detecting organ abnormalities, and identifying vascular or musculoskeletal problems, largely because it offers real‐time visualization without exposing patients to harmful ionizing radiation or requiring bulky, immobile machinery. 

A common approach to ultrasound tomography is full-waveform inversion~(FWI)~\cite{Physics-2025-Lin,virieux2017introduction,Dual-2023-Caradoc, guasch2020full, virieux2009overview}, yet in the transcranial setting, it is challenging both physically and numerically. \emph{Physically}, the skull--brain speed-of-sound contrast ($\sim\!3,000$\,m/s vs.\ $\sim\!1,520$\,m/s) induces strong attenuation, refraction, mode conversion, and phase aberration: (i) high-frequency components poorly penetrate bone, and (ii) surviving signals are distorted and low-SNR; moreover, full-aperture coverage around the head is clinically and manufacturing-wise infeasible, exacerbating limited-view artifacts. \emph{Numerically}, FWI minimizes a data–simulation misfit, but the objective is highly nonconvex and prone to cycle skipping without low-frequency content and a good initial model—both scarce in practice. Identifiability further degrades with band-limited sources, uncertainties in the source wavelet and array geometry, and modeling errors (inaccurate skull properties, attenuation/dispersion, elastic/anisotropic effects, boundary conditions), while multi-parameter inversions suffer parameter cross-talk. Computationally, each iteration requires many large-scale PDE and adjoint solves with heavy memory for checkpointing, rendering 3D problems expensive. Consequently, while ultrasound holds clear advantages in accessibility and safety over MRI and CT, its ability to deliver high-resolution, noninvasive brain images in real time is hampered by the skull’s acoustic properties and the limitations of current computational methods. Overcoming these barriers is critical to unlocking the full potential of ultrasound for rapid, point-of-care neurological assessment, particularly in time-sensitive scenarios such as stroke. In this work, we take a hybrid path, forming a physics-consistent method with machine learning to obtain quantitative SoS efficiently.

\begin{figure} [t]
\centering
\includegraphics[width=1.0\linewidth]{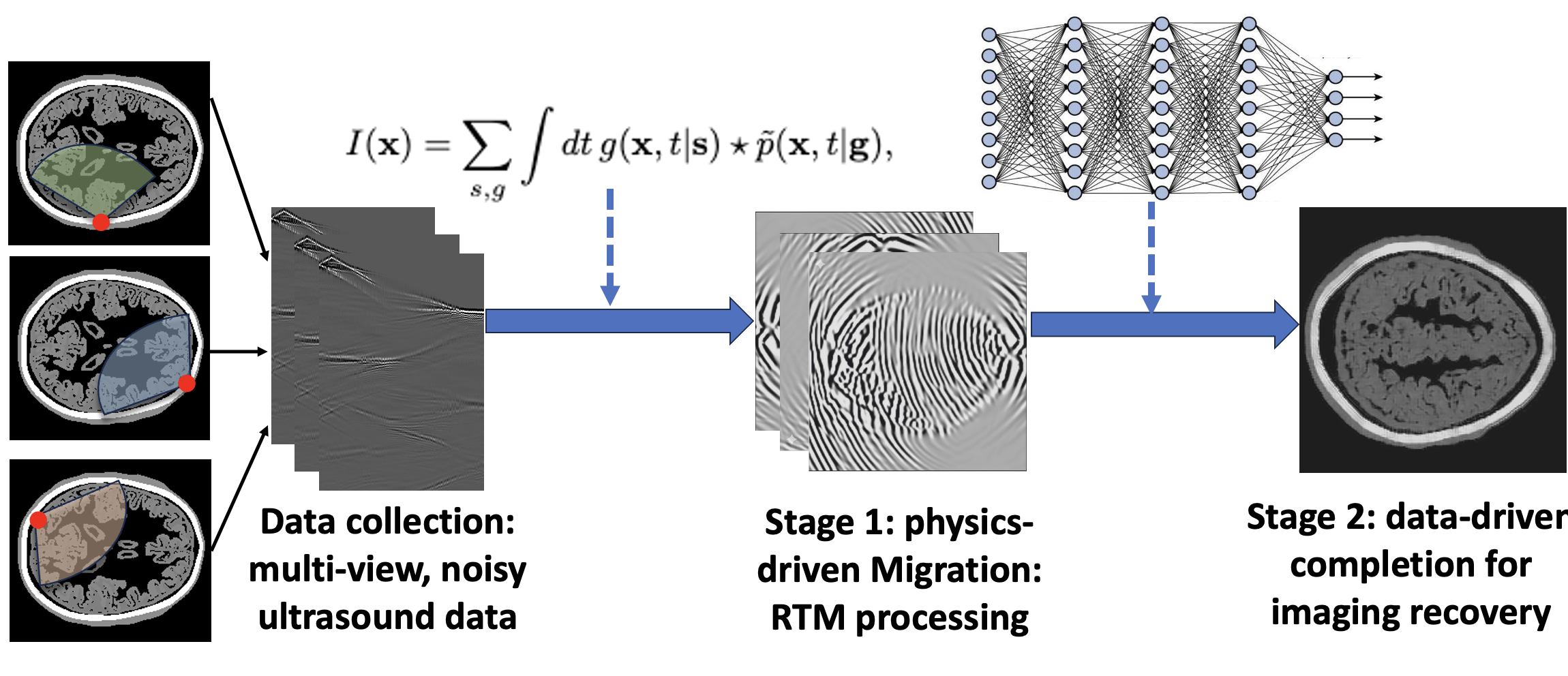}
\vspace{-.3in}
% \caption{The overview of the proposed workflow.\textcolor{red}{(this figure needs revision -- two stages)}} 
\caption{\textbf{BrainPuzzle workflow.} Stage~1: physics-driven RTM/TRA converts multi-view channel data into migration fragments under partial-aperture, low-SNR conditions. Stage~2: a transformer+GAU network fuses fragments and outputs a quantitative speed-of-sound (SoS) map. Movable partial-array sweeps provide multi-view coverage.}
\label{fig:workflow1}
\vspace{-.2in}
\end{figure}

In recent years, the geophysical community has made remarkable progress in addressing high speed of sound contrasts in seismic wave propagation using techniques such as reverse time migration (RTM)~\cite{baysal1983reverse,virieux2017introduction}. RTM were originally developed for hydrocarbon exploration in complex geological settings like salt basins, where the speed of sound in salt (4,500~m/s) significantly exceeds that in surrounding sedimentary layers~(2,500~m/s)~\cite{wang2019full, thiel2019comparison, privitera2016full, cruz2021tupi}. By accurately modeling wave propagation in these heterogeneous media, RTM has proven effective in reconstructing high-resolution subsurface images despite strong acoustic impedance mismatches. This same challenge of imaging through high speed of sound barriers arises in medical ultrasound applications—particularly in brain imaging—where the skull presents a substantial contrast in speed of sound relative to brain tissue. Similar to how salt structures obstruct seismic imaging of deeper formations, the skull impedes ultrasound wave transmission, complicating efforts to visualize brain structures. This parallel suggests that the success of RTM in geophysical imaging could inform advances in medical imaging, where it is recognized as the time reversal acoustics (TRA) technique.

While RTM/TRA has been highly successful in geophysics, a direct translation to transcranial ultrasound does not work out of the box. First, \emph{acoustic coupling} is problematic: large-aperture probes couple poorly to the rigid, curved skull, unlike compliant soft-tissue surfaces; immersive, water-coupled multi-transducer rigs have been proposed but are too complex and costly for routine clinical use. Second, \emph{skull penetration} is intrinsically hard at $\sim$2\,mm wavelength: high-frequency content is attenuated and phase-aberrated, calling for precise focusing (e.g., RTM/TRA) even under low SNR. Third, \emph{modeling fidelity} is limited by strong speed-of-sound contrast and uncertainty in bone properties, dispersion/attenuation, and bone–soft-tissue interfaces.  To address these constraints, we introduce \emph{BrainPuzzle}, a \textbf{hybrid} two-stage pipeline that separates \emph{physics formation} from \emph{learned completion}. \textbf{Stage~1} applies physics-driven RTM (TRA) to convert low-SNR, limited-aperture waveforms into \emph{migration fragments} that localize boundaries and preserve kinematics. \textbf{Stage~2} performs ML mapping from fragments to a coherent, quantitative SoS image using a transformer-based super-resolution network augmented with a graph-based attention unit (GAU) to reason over inter-fragment relations and suppress noisy or redundant inputs. For practicality, we adopt a \textbf{partial-array} acquisition—implemented as a movable, low-count transducer set that sweeps multiple views—reducing hardware cost and improving local coupling, while the two-stage algorithm compensates for the missing aperture (see overview in Fig.~\ref{fig:workflow1}).

In the following sections, we detail the data generation and acquisition setup, describe the two-stage \emph{BrainPuzzle} framework, and present experiments that validate its effectiveness for quantitative transcranial SoS reconstruction under limited-aperture, low-SNR conditions.

\begin{figure}
\centering
\vspace{-.2in}
\includegraphics[width=0.8\columnwidth]{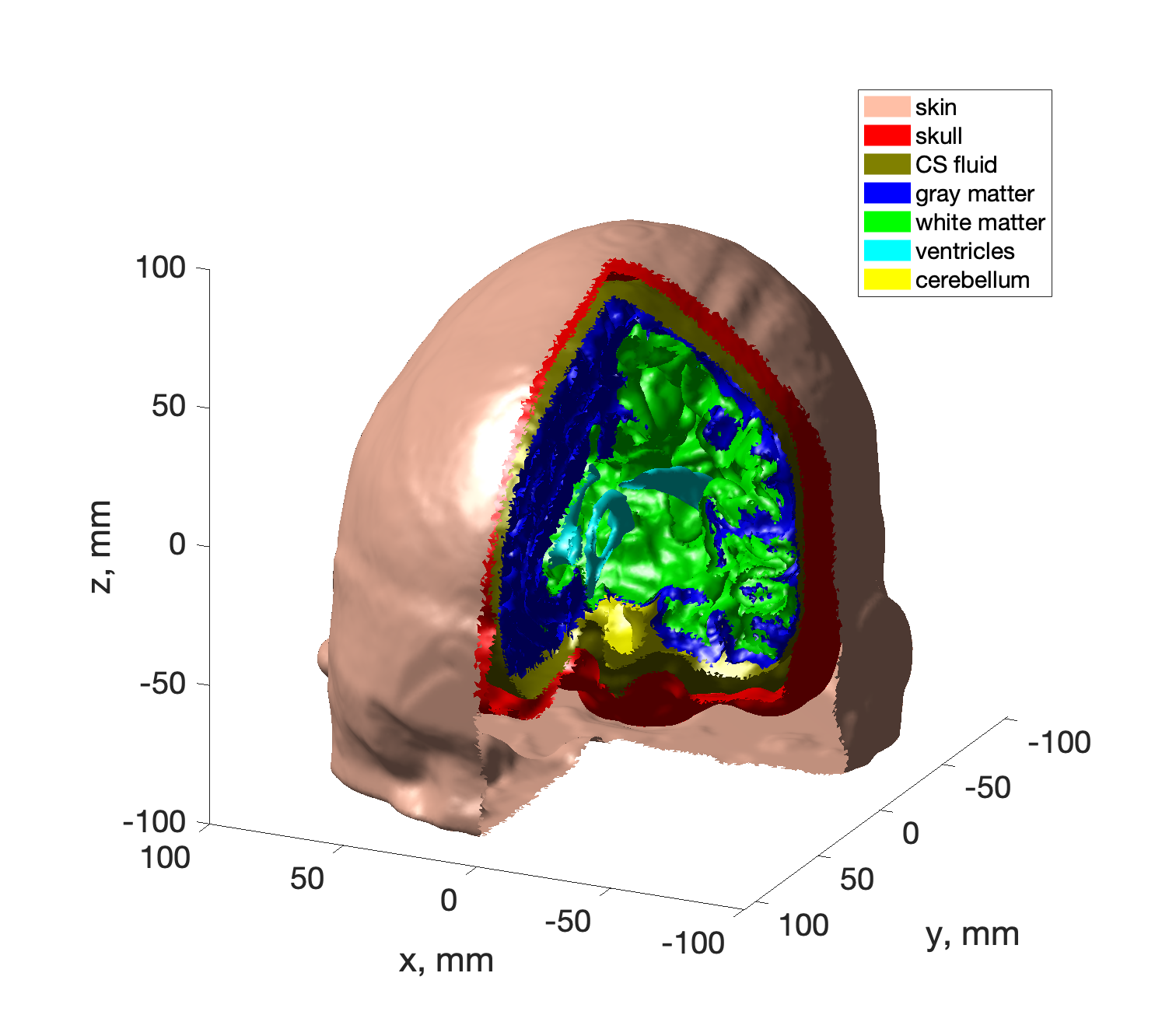}
% \caption{3D view of the skull model.~(\textcolor{red}{Axis is not easy to visualize.})}
\vspace{-.2in}
\caption{\textbf{3D rendering of the phantom.} Outer skin shown for context. Axial 2D slices are extracted along the $z$-axis for channel simulation and reconstruction.}
\label{fig:skull_3d_whole}
\vspace{-.2in}
\end{figure}

\section{Numerical Brain Phantom and Ultrasound Channel Data Generation}
\label{sec:dataset}

To the best of our knowledge, there is no openly available transcranial \emph{ultrasound channel} (raw RF) dataset for benchmarking. We therefore construct a 3D numerical speed-of-sound (SoS) phantom and synthesize channel data for evaluation. As shown in Figure~\ref{fig:skull_3d_whole}, the phantom comprises six tissue classes—skull, subarachnoid cerebrospinal fluid (CSF), cortical gray matter, subcortical white matter, ventricular CSF, and the cerebellum—illustrated in the supplementary file; their SoS values are listed in Table~\ref{table:skull_velocity}. From outer to inner across the cerebral convexity, the ordering is: \emph{skull}, \emph{subarachnoid CSF}, \emph{cortical gray matter}, \emph{subcortical white matter}, and \emph{ventricular CSF}. The \emph{cerebellum} is modeled as a separate posterior fossa structure rather than an inner cerebral layer.  As shown in Figure~\ref{fig:skull_3d_whole} and Table~\ref{table:skull_velocity}, accurate ultrasound reconstruction of the brain is inherently challenging due to the large acoustic contrast between soft tissues and the skull. Physically, this mismatch in impedance and velocity induces strong reflections, refraction, and mode conversion at bone–tissue interfaces, distorting transmitted wavefronts. The skull’s heterogeneous and anisotropic structure further introduces phase aberrations and multiple scattering, which blur spatial resolution and degrade image fidelity. Robust modeling and compensation of these effects are therefore essential for realistic transcranial ultrasound simulation and inversion.

% As observed in Table~\ref{table:skull_velocity}, the large variance of  the sound speed between soft tisues and skull creates challenge.... 

\begin{table}
\small
\centering
\vspace{-.2in}
\caption{Velocity Values of Numerical Brain Phantom}
\label{table:skull_velocity}
    \small
  \begin{adjustbox}{width=0.45\textwidth}
\begin{tabular}{|c|c|c|c|c|c|c|c|} 
\hline
                                                         & Skin & Skull & \begin{tabular}[c]{@{}c@{}}CS \\fluid\end{tabular} & \begin{tabular}[c]{@{}c@{}}Gray \\matter\end{tabular} & \begin{tabular}[c]{@{}c@{}}White \\matter\end{tabular} & Ventricles & Cerebellum  \\ 
\hline
\begin{tabular}[c]{@{}c@{}}Velocity\\~(m/s)\end{tabular} & 1,700 & 3,000  & 1,550                                               & 1,500                                                  & 1,480                                                   & 1,510       & 1,520        \\
\hline
\end{tabular}
\end{adjustbox}
\vspace{-.2in}
\end{table}

The focus of our work is 2D reconstruction. So, we extract axial slices by cutting the 3D model along the $z$-axis, yielding 150 sections of size $200\times221$ with in-plane grid spacing of 0.7\,mm. With the SoS phantom in place, we synthesize multi-view \emph{channel} (raw RF) data by solving the acoustic wave equation (Sec.~\ref{sec:physics} and Physical Imaging Methods) with a band-limited Ricker source centered at 300~kHz. We use a time step of $\Delta t=5\times10^{-8}$~s and record for 0.25~ms, yielding $T=5001$ time samples per trace.

\begin{figure*}

\centering
\vspace{-.2in}
\includegraphics[width=1.5\columnwidth]{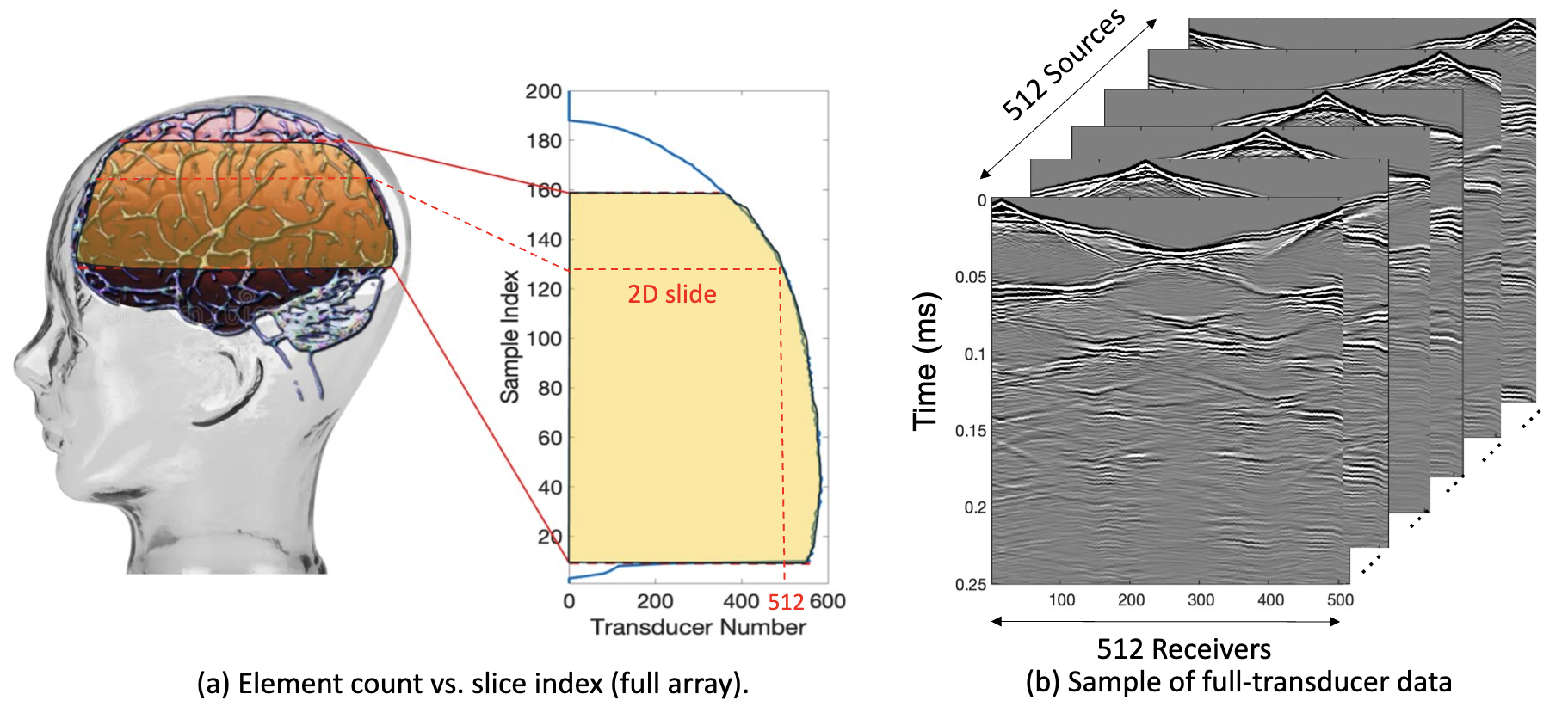}
% \caption{Transducer number variation across sample index. The yellow region indicates the range of sample indices selected for building the dataset.}
\vspace{-.2in}
\caption{(a) \textbf{Element count vs. slice index (full array).} Number of transducers $N$ varies with head circumference; yellow band marks slices used in the dataset (indices 10–159), with $N\in[358,585]$. (b) \textbf{Full-transducer channel data (2D slice with dashed red line at (a) ).} Shape $(T,N_s,N_r)=(5001,512,512)$ for time, sources, and receivers. One element transmits per shot; echoes are recorded on all 512 receivers. Sampling $\Delta t=5\times10^{-8}$\,s (0.25\,ms total). Here $N_s=N_r=512$ for this slice (varies across slices; see Fig.~\ref{fig:sample_num}).}
\label{fig:sample_num}
\vspace{-.2in}
\end{figure*}

\noindent\textbf{Full-transducer configuration (idealized).~}For each axial slice, we distribute point transducers along the head contour to approximate full-aperture coverage. Because skull circumference varies with slice depth, the number of elements $N$ changes across slices (Figure~\ref{fig:sample_num}a); we exclude very small top/bottom slices and retain indices 10–159 (yellow band in Figure~\ref{fig:sample_num}), where $N\in[358,585]$. We sequentially fire each element as a source while recording on all others, producing a data tensor $\mathbf{D}\in\mathbb{R}^{T\times N_s\times N_r}$ with $T=5001$ time steps and $N_s=N_r=N$. Thus, the maximum per-slice dimension is $(5001,585,585)$; Figure~\ref{fig:sample_num}b shows a representative case with $(5001,512,512)$. We refer to this collection as the \emph{full-transducer} dataset.

Although informative for benchmarking, a dense, simultaneous full-aperture array is impractical: each element requires dedicated drive/receive electronics and high-throughput acquisition; data volume and processing cost scale quadratically with $N$; and dense rigid arrays couple poorly to the curved skull, impacting comfort and SNR. These constraints motivate our \emph{partial-array} strategy, where a movable, low-count set sweeps multiple views; the hybrid BrainPuzzle pipeline compensates algorithmically for the missing aperture.

\section{Related Work}

\subsection{The applications of RTM and FWI}

Time reversal acoustics~(TRA) \cite{fink2000time} and full waveform inversion~(FWI) \cite{guasch2020full} are physics-based ultrasound imaging methods that have demonstrated significant potential in medical applications. TRA has been successfully applied for focusing and compensation through heterogeneous tissues, while FWI, originally developed in geophysics, has recently been adapted for ultrasound to reconstruct tissue properties with high resolution. Together, these methods illustrate the promise of physics-informed approaches for advancing both diagnostic and therapeutic imaging.

TRA utilizes the recorded ultrasound signal, reverses it in time, and retransmits it back into the medium. This process effectively refocuses the wave energy to its source location. Widely applied in medical imaging, this technique is instrumental in accurately localizing anomalies such as tumors or lesions. By utilizing the full wavefield information from ultrasound signals, FWI iteratively refines a model of the medium's internal structure. This advanced technique excels at reconstructing detailed acoustic speed of sound maps, effectively highlighting variations in tissue density, stiffness, and other key properties. Both of them have been adapted for medical ultrasound imaging to non‐invasively characterize human organs, such as the brain~\cite{Ultrasound-2024-Li, Truncated-2024-Wu, Amortized-2023-Orozco, Dual-2023-Caradoc, guasch2020full}, breast~\cite{Learned-2023-Lozenski, fan2022model, roy2016ultrasound,Waveform-2015-Wang, agudo20183d}, and other tissues~\cite{Openpros-2025-Wang, New-2024-Pan, wang2015reverse}, demonstrating their versatility and potential in complex diagnostic applications. Notably, the recent study by Orozco et al.~\cite{Amortized-2023-Orozco} introduced a physics-informed summary statistic framework for brain imaging, illustrating how TRA can be leveraged to significantly reduce data dimensionality while preserving essential physical information. For a comprehensive overview of wave-based imaging techniques and their broad medical applications, readers are referred to the survey in~\cite{Physics-2025-Lin}.

\subsection{Transformer in image generation}
Transformers~\cite{parmar2018image}, initially designed for language processing, have been innovatively adapted for image generation by treating images as sequences of pixels or patches, similar to how they treat text. The core feature of Transformer is the self-attention mechanism, which allows it to process the entire image at once, capturing complex relationships and patterns within the visual data. This global perspective is key to generating high-quality, contextually coherent images. Transformers are trained on large image datasets, enabling them to either create new images from scratch or modify existing ones based on learned visual representations. Their application in image generation spans from creating photo-realistic images to artistic designs, benefiting significantly from their ability to understand and manipulate complex scenes. In a recent survey~\cite{han2022survey}, it is indicated that the Transformer has already shown tremendous success in several vision problems such as image super-resolution~\cite{lu2022transformer, yang2020learning, li2022hst}, image translation~\cite{kim2022instaformer, torbunov2023uvcgan, zheng2022ittr}, image segmentation~\cite{gao2021utnet, huang2021missformer, jain2023oneformer}, etc. Furthermore, the concept of a Transformer has demonstrated promising results in addressing medical imaging problems in a recent survey~\cite {shamshad2023transformers}.

\subsection{Single image super-resolution (SISR)}
Single image super-resolution (SISR) via deep learning has been the subject of many investigations in computer vision. These methods derive their power primarily from the utilization of convolutional network layers~\cite{albawi2017understanding}, which extract spatial texture features and transform them through complex non-linear mappings to recover high-resolution data. One of the earliest SR methods for SISR is SRCNN~\cite{dong2014learning}, which learns an end-to-end mapping between coarse-resolution and high-resolution images by employing a series of convolutional layers. Another scheme is the skip-connection layers~\cite{Duong2021, Dai2019, zhang2018residual, ahn2018fast, Tai2017},  which enable the bypassing of abundant low-frequency information and emphasize the relevant information to improve the stability of the optimization process in deep neural networks. 
Several investigators have explored the adversarial training objective by using the generative adversarial network (GAN) for  SISR~\cite{ledig2017photo, chen2017fsrnet, wang2018recovering, wang2018esrgan, karras2018progressive, gan8759375, cheng2021mfagan, Long2021}. After that, the Transformer~\cite{parmar2018image, fang2022cross,yang2020learning, lu2022transformer,  fang2022hybrid, wang2022detail} and diffusion model~\cite{gao2023implicit, li2022srdiff, moser2024diffusion} have also been introduced into the SISR problem.  

\subsection{Applications of ultrasound computed tomography}
Ultrasound computed tomography (USCT) with full‐waveform inversion (FWI) is a promising method for medical imaging that uses the physics of wave propagation to create detailed maps of tissue properties~\cite{lozenski2024learned, zhang2024cross}. In this approach, multiple ultrasound transducers are placed around the patient to capture how sound waves travel through and reflect off different tissues. An iterative algorithm then refines an initial model of the anatomy until the simulated wave patterns match the actual measurements. This process can yield higher resolution images than conventional ultrasound, helping clinicians diagnose and monitor conditions such as breast tumors~\cite{vargasbreast}, liver diseases~\cite{sharon2024real}, lung diseases~\cite{goyal2025ultrasound}, and brain disorders~\cite{Dual-2023-Caradoc, marty2021acoustoelastic, guasch2020full}. Although USCT with the FWI technique can be computationally demanding, ongoing improvements in computing power are making it increasingly feasible in clinical settings. By providing a non‐ionizing, portable, and potentially lower‐cost alternative to MRI and CT, USCT with FWI holds the potential to expand the use of ultrasound across a range of medical applications.

\section{Physical Imaging Methods}
\label{sec:physics}

In this section, we will introduce the theory of TRA and FWI. Both physical imaging methods utilize full-waveform ultrasound data to generate images that reveal detailed brain structures.

\subsection{Forward modeling}
The ultrasound data is generated with the acoustic wave equation \begin{equation}
\nabla^2d-\frac{1}{{c}(\mathbf{x})^2}\frac{\partial ^2d}{\partial t ^2}=s(\mathbf{x},t),
\label{eq:Acoustic}
\end{equation}
where $\nabla^2$ is the Laplace operator, $c$ represents the speed of sound~(SoS), $d$ is the pressure field and $s(\mathbf{x},t)$ is source term. The SoS, ${c}(\mathbf{x})$, depends on the spatial location $\mathbf{x}$ while the pressure field $d$ and source term $s(\mathbf{x},t)$ depend on the spatial location and time $(\mathbf{x},t)$. The finite-difference method is employed to simulate the ultrasound wave propagation.

\subsection{Full waveform inversion}
% \begin{thebibliography}{5}
Full waveform inversion~(FWI)~\cite{virieux2009overview} reconstructs SoS by minimizing the difference between observed and simulated pressure fields using the waveform misfit function
\begin{equation}
\zeta = \frac{1}{2} \sum_{\mathbf{g},t|\mathbf{s}} [\Delta{d}(\mathbf{g},t|\mathbf{s}) ]^2,
\label{eq:gloss}
\end{equation}
where $\Delta{d}(\mathbf{g},t|\mathbf{s})={d}_{\mathrm{cal}}(\mathbf{g},t|\mathbf{s}) - {d}_{\mathrm{obs}}(\mathbf{g},t|\mathbf{s})$ is the waveform residual.
$\mathbf{s}$ and $\mathbf{g}$ are the sources and receivers.

%The steepest descent method gives
%\begin{equation}
% {{m}(\mathbf{x})}^{(i+1)}= {{m}(\mathbf{x})}^{(i)}-\alpha\sum_{s,g}\Delta{d}(\mathbf{g},t|\mathbf{s})\frac{
% \partial{{d}_{cal}(\mathbf{g},t|\mathbf{s})}}{\partial{{m}(\mathbf{x})}},
% \label{eq:itrative}
%\end{equation} 
%where $\alpha$ is the step length and $i$ is the iteration index.

The SoS ${c}(\mathbf{x})$ can be estimated using any gradient-based method. The gradient of misfit function $\zeta$ with respect to the SoS $c(\mathbf{x})$ is
\begin{eqnarray}
\frac{\partial{\zeta}}{\partial{c(\mathbf{x})}}&=
\frac{1}{c^3(\mathbf{x})}\sum_{s,g}\int{dt}
\dot{{g}}(\mathbf{x},t|\mathbf{s})\star
\dot{{p}}(\mathbf{x},t|\mathbf{g}),
 \label{eq:Frechet}
\end{eqnarray}
where the dot indicates the time differentiation, ${{g}}(\mathbf{x},t|\mathbf{g})$ is the forward wavefield 

\begin{eqnarray}
{{g}}(\mathbf{x},t|\mathbf{s})={{G}}(\mathbf{x},t|\mathbf{s})\star{w(t)},\label{eq:forward}
\end{eqnarray}
where $w(t)$ is the source wavelet and ${{G}}(\mathbf{x},t|\mathbf{s})$ is the Green's function associated with the acoustic wave equation~\eqref{eq:Acoustic}. The term ${p}(\mathbf{x},t|\mathbf{g})$ is the wavefield computed by backprojecting the waveform residual $\Delta{d}(\mathbf{g},t|\mathbf{s})$
\begin{eqnarray}
{{p}}(\mathbf{x},t|\mathbf{g})={{G}}(\mathbf{x},-t|\mathbf{g},0)\star{\Delta{d}}(\mathbf{g},t|\mathbf{s}),\label{eq:back}
\end{eqnarray}

The SoS gradient is calculated by the crosscorrelation of the forward-propagated ${{g}}(\mathbf{x},t|\mathbf{s})$ and the backward-propagated wavefield ${{p}}(\mathbf{x},t|\mathbf{g},0)$ of the residual $\Delta{d}(\mathbf{g},t|\mathbf{s})$. ~

\subsection{Time reversal acoustics}
Time reversal acoustic~(TRA)~\cite{fink1992time} makes use of the time-reversibility of the wave equation. By recording acoustic waves at an array of sensors, time-reversing these signals, and re-emitting them into the medium, the waves retrace their paths and converge back to the reflection point. This method focuses energy precisely at the reflection location, which reveals accurately localizing anomalies such as tumors
or lesions.

Instead of backpropagating the waveform residual $\Delta{d}$, TRA uses the full recorded data $d_{\mathrm{obs}}(\mathbf{g},t|\mathbf{s})$ during backpropagation, resulting in an image that represents the energy focusing at the original reflection location
\begin{eqnarray}
I(\mathbf{x})&=
\sum_{s,g}\int{dt}
{{g}}(\mathbf{x},t|\mathbf{s})\star
\tilde{{p}}(\mathbf{x},t|\mathbf{g}),
 \label{eq:tra}
\end{eqnarray}
where $\tilde{{p}}(\mathbf{x},t|\mathbf{g})$ is computed by backprojecting the recorded data $d_{\mathrm{obs}}(\mathbf{g},t|\mathbf{s})$ defined as
\begin{eqnarray}
{\tilde{p}}(\mathbf{x},t|\mathbf{g})={{G}}(\mathbf{x},-t|\mathbf{g},0)\star{{d}_{obs}}(\mathbf{g},t|\mathbf{s}).\label{eq:back2}
\end{eqnarray}

\begin{figure*} [!t] 
\vspace{-.2in}
%\vspace{-.2in}
\centering
\includegraphics[width=0.8\textwidth%,height=0.27\textwidth
]{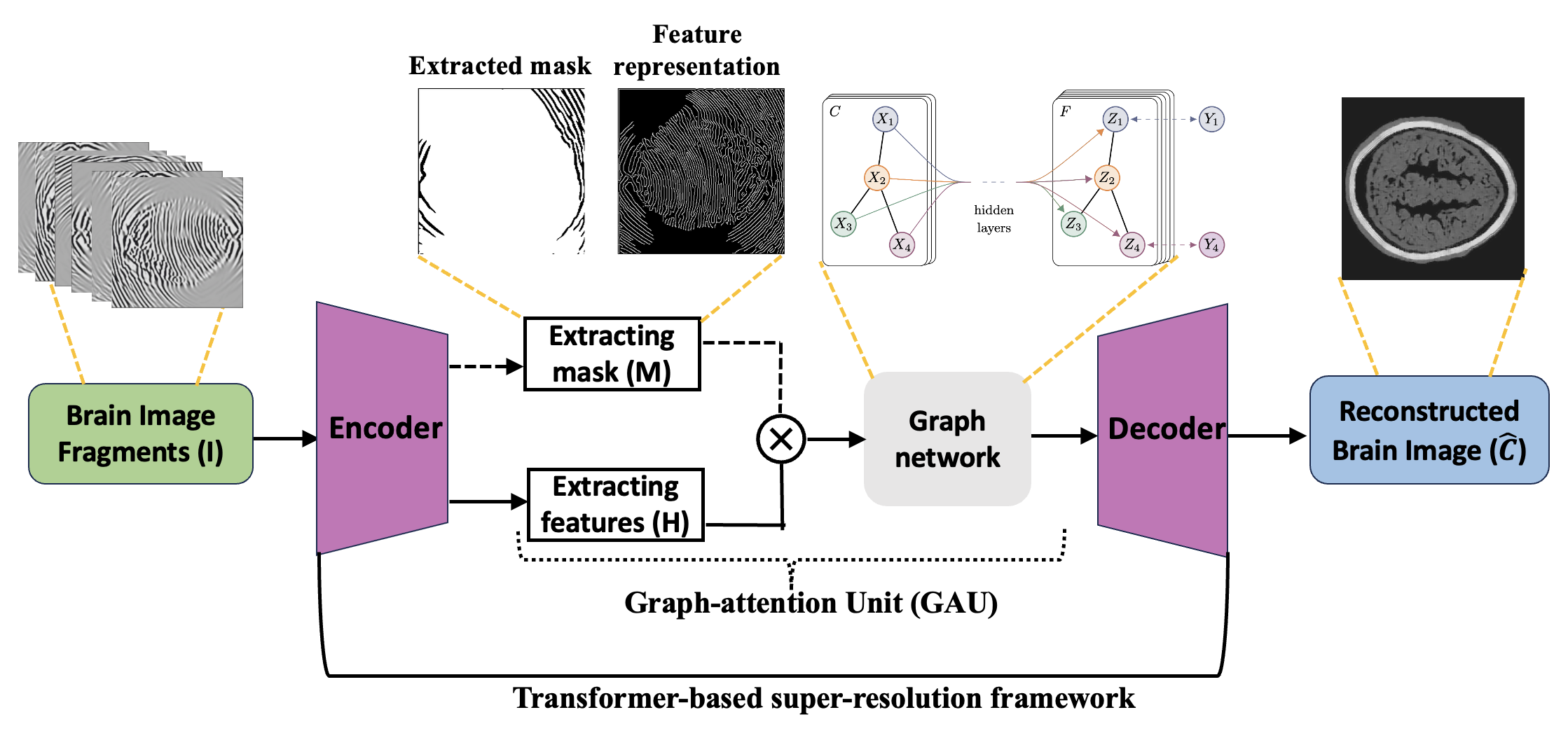}
%\vspace{-0.1in}
\vspace{-.2in}
\caption{\textbf{Stage-2 network of \emph{BrainPuzzle}.} Multi-view TRA fragments are embedded (with positional and view encodings) and passed through a shared transformer encoder; a lightweight super-resolution upsampler restores spatial detail; a graph-based attention unit (GAU) performs cross-fragment fusion and adaptive weighting; a decoder and reconstruction head output the quantitative SoS map. The architecture naturally fuses partial-array acquisitions by aggregating fragments from multiple probe positions.}
\vspace{-.2in}
\label{fig:overall_struc}
\end{figure*}

\subsection{FWI vs TRA}
TRA produces a reflectivity image that offers a clear depiction of structural boundaries. It enables fast, real-time imaging with high spatial resolution, but it is limited in its ability to quantify physical properties.

In contrast, FWI generates a SoS image of the brain, providing quantitative information that is valuable for detecting pathologies not visible in reflectivity images. However, FWI requires more complex data processing and depends on having a good initial model, which makes it less popular in clinical settings.

Machine learning offers the potential to combine the advantages of both methods. By leveraging reflectivity images along with the SoS information, neural networks can reconstruct high-quality SoS from reflectivity data~\cite{zhang2021deep,zhang2021least,geng2022deep}. Thus, in our methodology, instead of using brain fragments to reconstruct a full reflectivity image, we directly focus on reconstructing a high-quality speed of sound image.

\section{Methodology: Two-Stage Hybrid Reconstruction}

Our proposed \emph{BrainPuzzle} (Figure~\ref{fig:workflow1}) comprises a physics stage that forms localized \emph{TRA fragments} and a learned stage that assembles a coherent, quantitative SoS image.

\subsection{Stage 1: Physics-Driven TRA}
% Given multi-view recordings $d_{\mathrm{obs}}(\mathbf{g},t|\mathbf{s})$, we compute a time-reversal (RTM) image via Eq.~(\ref{eq:tra}). TRA yields \emph{reflectivity-like fragments} that encode boundary geometry and partial kinematics even under weak signals and sparse views; however, they remain spatially disjoint with limited aperture.

In the transcranial regime, raw channel data are low-SNR and strongly phase-aberrated, and the aperture is incomplete. Reverse time migration (time-reversal acoustics, TRA) is attractive because it (i) performs \emph{coherent backpropagation} and stacking of the full recorded wavefield, which improves effective SNR through phase-coherent summation; (ii) is \emph{model-light}—it relies on a smooth background (or nominal) model rather than a high-fidelity initial SoS, and thus avoids the strong initialization sensitivity of FWI; (iii) preserves \emph{kinematic fidelity} (boundary locations and travel times) even when amplitudes are not quantitatively reliable; and (iv) is \emph{computationally efficient}, requiring a single forward/backward propagation per source without inner optimization loops.

\noindent \textbf{Imaging operator.} Given multi-view recordings $d_{\mathrm{obs}}(\mathbf{g},t|\mathbf{s})$, we form a time-reversal image via Eq.~(\ref{eq:tra}):
\begin{equation}
I(\mathbf{x})=\sum_{s,g}\int dt\, g(\mathbf{x},t|\mathbf{s}) \star \tilde{p}(\mathbf{x},t|\mathbf{g}),
\label{eq:rtm_method}
\end{equation}
with $g(\mathbf{x},t|\mathbf{s})=G(\mathbf{x},t|\mathbf{s})\!\star\!w(t)$ and $\tilde{p}(\mathbf{x},t|\mathbf{g})=G(\mathbf{x},-t|\mathbf{g},0)\!\star\!d_{\mathrm{obs}}(\mathbf{g},t|\mathbf{s})$. The cross-correlation (imaging condition) focuses energy at reflectors where forward and backpropagated wavefields are phase-aligned, yielding \emph{reflectivity-like} maps that encode boundary geometry and partial kinematics under limited aperture and low SNR.

\noindent \textbf{From TRA images to fragments.} For each probe placement (or sweep) $k$, we compute an TRA image $I_k(\mathbf{x})$. Because each limited-aperture view illuminates only part of the anatomy and suffers from view-dependent aberration, the individual $I_k$ is \emph{informative but incomplete}. We apply mild, view-wise normalization (e.g., trace balancing and bandpass consistent with the source) to reduce illumination bias and collect the set
\[
\mathcal{I}=\{\,I_k(\mathbf{x})\,\}_{k=1}^{n},
\]
which we term \emph{TRA fragments}. These fragments (i) localize interfaces with higher stability than raw RF, (ii) provide complementary coverage across views, and (iii) raise the effective SNR via coherent stacking—yet they remain spatially disjoint and amplitude-inconsistent due to missing angles and skull-induced aberration.

TRA thus serves as a \emph{physics front-end} that converts challenging, low-SNR channel data into structured image-space evidence with reliable kinematics. Stage~2 of \emph{BrainPuzzle} then learns the residual tasks—completion under limited aperture and mapping from reflectivity-like fragments to quantitative SoS—where data-driven priors are most effective.

\subsection{Stage 2: ML Mapping from TRA to SoS}

Given the set of TRA fragments $\mathcal{I}=\{I_k(\mathbf{x})\}_{k=1}^{n}$ from Stage~1, Stage~2 learns a mapping $f_\theta:\mathcal{I}\rightarrow \hat{\mathbf{c}}$ that produces a quantitative SoS image $\hat{\mathbf{c}}$. This choice is motivated by physics: although TRA is a reflectivity-style imaging operator, its outputs are \emph{shaped} by the underlying SoS through travel times and focusing; hence, fragments encode reliable kinematic cues (interfaces, arrival geometry) with higher effective SNR than raw RF. The learning task is therefore reduced to \emph{completion and calibration}—inferring missing views and converting kinematics into quantitative SoS—rather than relearning wave physics end-to-end.

BrainPuzzle adopts a multi-view transformer encoder–decoder with super-resolution and a graph-based attention unit (GAU) for fusion (Figure~\ref{fig:overall_struc}). Each fragment $I_k$ is embedded and passed through a \emph{shared} encoder to extract multi-scale features; an upsampling module restores high-frequency detail; GAU performs cross-fragment reasoning and adaptive weighting; finally, a decoder produces the dense SoS field $\hat{\mathbf{c}}$.

\noindent \textbf{Fragment embedding, encoder, and super-resolution.~}Each fragment $I_k$ is first mapped to a feature map $e_k$ by a shallow stem (conv+norm+nonlinearity) and augmented with
\emph{positional} and \emph{view} embeddings to retain spatial coordinates and view identity. A \emph{shared} transformer encoder then processes $\{e_k\}$, using multi-head self-attention within each fragment to capture long-range dependencies and intra-fragment structure, producing latent features $\{h_k\}$. To recover fine detail, a lightweight upsampler~(transposed convolutions or interpolation followed by convolution) transforms $\{h_k\}$ into high-resolution features $\{r_k\}$, increasing spatial resolution while sharpening edges and textures critical for accurate SoS estimation.

\noindent \textbf{Graph-based attention fusion (GAU).~}The graph-based attention unit (GAU) is designed to refine high-resolution fragment features $\textbf{r} = \{r_1, r_2, \dots, r_n\}$ by modeling spatial relationships and selectively emphasizing informative content, which is critical for accurate brain image reconstruction from limited ultrasound fragments. GAU comprises two key components: a graph convolutional network (GCN) that captures spatial dependencies among fragments, and an attention mechanism that adaptively weighs each fragment’s importance in the reconstruction process.

To capture inter-fragment dependencies, GAU represents the input fragments as nodes in a fully connected graph, where the edges reflect learned similarities. These relationships are encoded in a learnable adjacency matrix $\textbf{A} \in \mathbb{R}^{n \times n}$, where $A_{ij}$ quantifies the influence of fragment $j$ on fragment $i$. Initially, $\textbf{A}$ may be initialized based on feature similarity (e.g., cosine similarity), but it is updated during training to better represent spatial and semantic correlations among fragments.

The GCN processes the feature set $\textbf{r}$ using this adjacency matrix, enabling each node to aggregate information from its neighbors. This message-passing mechanism~\cite{xu2024utilizing} allows the model to enhance each fragment’s representation by integrating contextual knowledge from related fragments. As a result, even if some fragments are missing or noisy, the GCN can infer missing content based on patterns learned across the graph, supporting robust and context-aware reconstruction.

In parallel, GAU incorporates an attention mechanism that assesses the relative importance of each fragment feature. Specifically, it generates a set of importance masks $\textbf{m} = \{m_1, m_2, \dots, m_n\}$, where each $m_i \in \mathbb{R}^d$ corresponds to a feature-wise weight vector applied to fragment feature $r_i$. These masks are generated using a lightweight convolutional subnetwork that shares structure with the encoder and ends with a sigmoid activation function. To normalize the attention weights across fragments, a softmax function is applied: $\hat{\textbf{m}} = \text{Softmax}(\textbf{m}),$ where each $\hat{m}_i$ assigns a learnable feature-wise weight to the corresponding $r_i$. Unlike global attention vectors, these masks operate on a per-fragment, per-feature basis, allowing the model to selectively amplify or suppress specific dimensions of each fragment based on its local importance. The refined high-resolution features are computed by element-wise multiplication: $\textbf{r} = \textbf{r} \cdot \hat{\textbf{m}},$ which modulates each fragment according to its learned importance. This allows the model to focus more on fragments that carry unique or critical information, enhancing reconstruction accuracy in scenarios with incomplete input.

In addition, GAU is integrated into the overall transformer-based super-resolution framework, forming a two-stage refinement pipeline. The transformer-based super-resolution framework focuses on enhancing each fragment by improving spatial resolution and restoring fine-grained texture using transformer-based self-attention and upsampling. Once fragments are enhanced, the GAU performs global reasoning by modeling inter-fragment relationships (via the GCN) and adaptively weighting their contributions (via the attention mechanism).

This complementary design allows the model to first ensure that each fragment is of high quality and then integrate them coherently based on their relevance and relationships. Together, these enable the BrainPuzzle framework to effectively reconstruct complete and detailed brain images from limited and fragmented ultrasound inputs.

\noindent \textbf{Decoder and loss function.~}A transformer decoder with skip connections (from the encoder) integrates the fused features and outputs the SoS estimate $\hat{\mathbf{c}}$ through a shallow reconstruction head. Skip connections preserve low-level spatial cues, while attention focuses on fragments most relevant to each spatial region.

We optimize a reconstruction loss that balances quantitative fidelity and structural realism:
\begin{equation}
\mathcal{L}_{\text{Recon}} = \underbrace{\mathrm{MSE}(\mathbf{c},\hat{\mathbf{c}})}_{\text{pixel fidelity}}
+ \alpha_{\text{Perc}}\, \underbrace{\|\phi(\mathbf{c})-\phi(\hat{\mathbf{c}})\|_2^2}_{\text{perceptual}},
\label{eq:Recons_loss}
\end{equation}
where $\phi(\cdot)$ extracts feature maps from a fixed VGG network. Operating on TRA fragments improves robustness to low SNR
and partial coverage, enhances sample efficiency versus raw-RF learning, and naturally accommodates the \textbf{partial-array}
acquisition by integrating movable views in GAU.

\section{Experiment}
\subsection{Experimental designs}
\subsubsection{Dataset}

\begin{figure} [t]
\vspace{-.2in}
\centering
\includegraphics[width=0.4\linewidth]{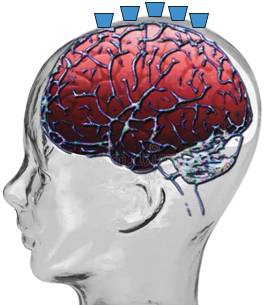}
\vspace{-.1in}
\caption{\textbf{Partial-array acquisition schematic.} A movable low-count transducer set (e.g., 50 elements) is placed at successive head positions to collect multi-view channel data. For each view, one element transmits while the others receive; repeating across sweeps yields partial-aperture coverage with improved coupling and reduced hardware cost.} 
\vspace{-.25in}
\label{fig:brain_model_partial}
\end{figure}

\begin{figure*}
\vspace{-.2in}
\centering
\includegraphics[width=1.6\columnwidth]{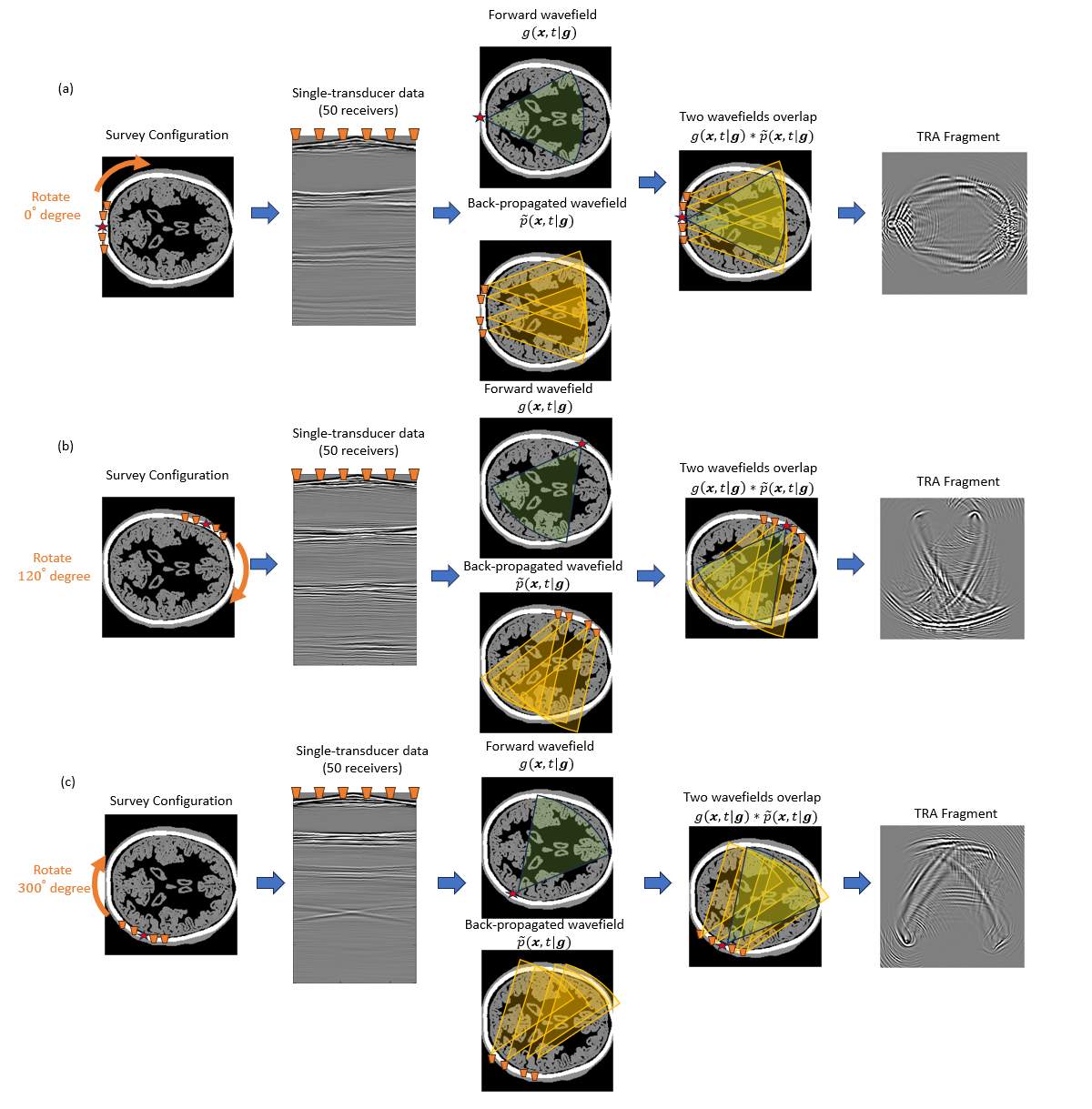}
% \caption{\textbf{Partial-Transducer Dataset: TRA Imaging with Rotating Single-Transducer Around the Brain.} A linear array with 51-element performs surveys by rotating around the head with single central transducer as source, other remaining 50 as receivers. (a) the array at rotation 0 $^\circ$, (b) the array at rotation 120 $^\circ$ and (c) the survey at rotation 300 $^\circ$. Each survey produces data with dimensions $(T,N_s,N_r)=(5001,10,50)$, corresponding to time, sources (sweeps), and receivers. TRA fragments are then generated through reverse time migration imaging.}
\vspace{-.2in}
\caption{\textbf{Partial-transducer dataset: TRA imaging with a rotating single-transducer array around the head.} A 51-element linear array performs surveys by rotating around the head, where the central transducer acts as the source and the remaining 50 elements serve as receivers. (a) Array position at 120$^\circ$, and (b) at 300$^\circ$. Each survey produces data with dimensions $(T, N_s, N_r) = (5001, 10, 50)$, corresponding to time, sources (sweeps), and receivers. The TRA fragments are subsequently generated through reverse time migration imaging.}
\label{fig:data_example_partial_new1}
\vspace{-.2in}
\end{figure*}

\begin{figure} [!t] 
\vspace{-.1in}
\centering
\subfigure[Partial-transducer dataset]{ \label{fig:a}{}
\includegraphics[width=0.35\columnwidth, height=0.32\columnwidth]{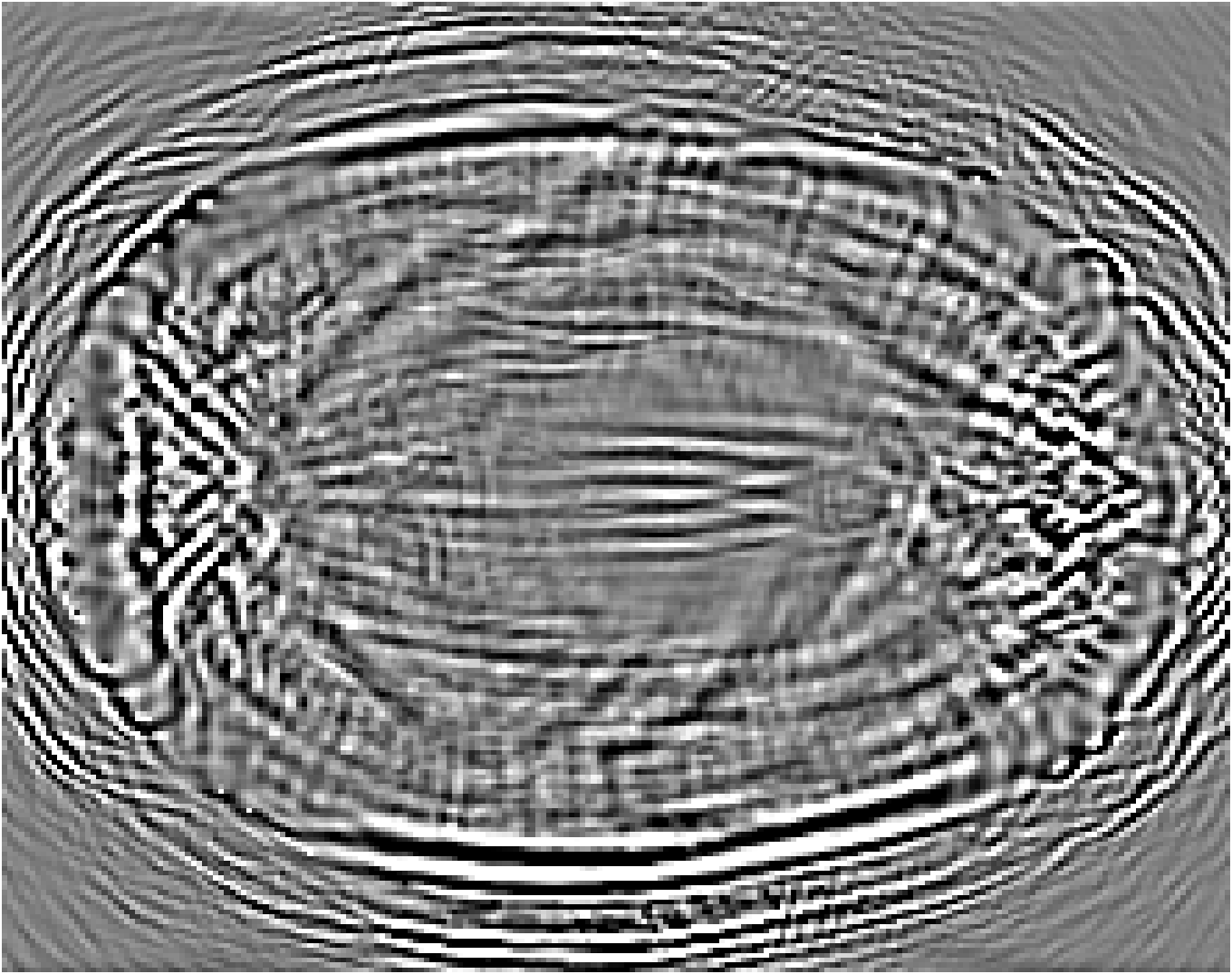}
}\hspace{1mm}
\subfigure[Full-transducer dataset]{ \label{fig:b}{}
\includegraphics[width=0.35\columnwidth, height=0.32\columnwidth]{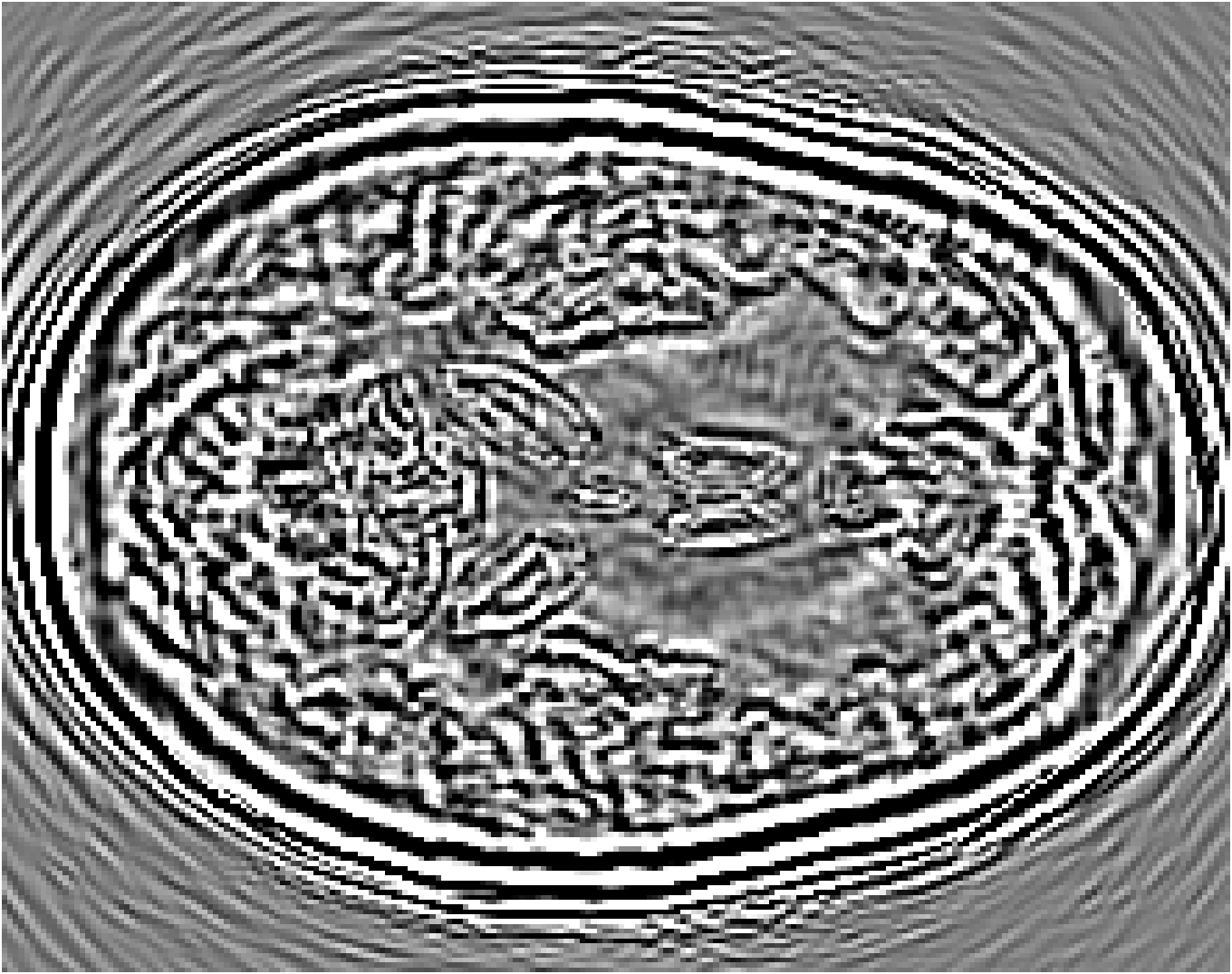}
}
\vspace{-.1in}
\caption{\textbf{Stacked TRA Image} with (a) Partial-transducer dataset and (b) Full-transducer dataset }
\label{fig:stack_rtm}
\vspace{-.2in}
\end{figure}

To evaluate performance, we construct two acquisition setups using the same horizontal 2D slice of the phantom: (i) an idealized full-transducer dataset (Figure \ref{fig:sample_num}(b)) providing near-uniform coverage along the head contour, and (ii) a partial-transducer dataset (Figure \ref{fig:data_example_partial_new1}) designed to reflect practical acquisition constraints. In the partial setup, a linear array with 51-element is repositioned around the head; for each view, the central element transmits and all other 50 elements receive. The partial setup includes 50 sweeps covering the full 360$^\circ$, meaning that for each sweep, the source and receivers rotate by $360/50 = 7.2^\circ$. Sweeping 50 views yields, per 2D slice, a channel tensor of shape $(T,N_s,N_r)=(5001,50,50)$.  The key distinction between these two datasets is their aperture: full-transducer provides near 360$^\circ$ coverage in a single placement, whereas partial-transducer attains coverage by aggregating neastest 50 receivers in a single placement, which is only around 36$^\circ$.
Particularly, the partial-transducer setup includes only 50 sweeps, while the full-transducer setup uses all receivers covering the entire brain, with each transducer acting as a source in turn while the others serve as receivers. Compared to the full-transducer, the partial-transducer achieves coverage by combining a much smaller number of views. For each sweep, we generate the forward wavefield ${{g}}(\mathbf{x},t|\mathbf{g})$ and the back-propagated wavefield $\tilde{{p}}(\mathbf{x},t|\mathbf{g})$ and generate the TRA fragment with Eq.~\eqref{eq:back2}~(example in Figure~\ref{fig:data_example_partial_new1}).
With the physical TRA method, the TRA fragment can be stacked to give the stacked TRA image. To validate the differences between the two datasets, we generated stacked TRA images using both datasets, as shown in Figure~\ref{fig:stack_rtm}. As shown in Figure~\ref{fig:stack_rtm}(b), the full-transducer dataset provides significantly greater detail in internal structures. Despite higher levels of noise and artifacts caused by the limited number of views, the partial-transducer dataset is still capable of capturing some meaningful tissue structures.

%The full-transducer dataset provides detailed internal structures, whereas the partial-transducer dataset fails to do so due to the limited number of views. 

\subsubsection{Experimental settings}

Each of the full-transducer and partial-transducer datasets contains 150 axial SoS slices, together with their corresponding TRA fragments. To ensure fair and consistent evaluation, 100 slices are randomly chosen for training and 50 for testing. Specifically, the performance of BrainPuzzle is evaluated by comparing it against several baseline models commonly used in medical imaging research. For this comparison, we implement BrainPuzzle alongside a widely recognized CNN-based baseline, U-Net~\cite{barkau1996unet}. In addition, we include two popular super-resolution models, RCAN~\cite{zhang2018image_backup} and CycGAN~\cite{chu2017cyclegan}, as well as two generative models: the transformer-based ViT~\cite{han2022survey} and the diffusion model-based LDM~\cite{moser2024diffusion}, as baseline methods. To further evaluate the contribution of each component of BrainPuzzle, we introduce two additional ablation baselines, $\text{BrainPuzzle}_\text{T}$, $\text{BrainPuzzle}_\text{G}$. Specifically, 
$\text{BrainPuzzle}_\text{T}$ is a transformer-based super-resolution model without including GAU. $\text{BrainPuzzle}_\text{G}$ is designed without the automatic attention mechanism within the GAU, thereby assessing the significance of this feature in the model's overall performance. In addition, we also include the physical method, FWI, in our comparisons. 
%Our dataset and implementation of the proposed model have been released\footnote{Dataset and model implementation: \url{https://drive.google.com/drive/folders/1V8limYC0-UJm-PbZWvwyfiReVdWHaFTS?usp=sharing}}

By including these diverse baselines and ablation studies, we aim to thoroughly analyze BrainPuzzle's effectiveness and identify the specific contributions of its core components to its superior performance in brain image reconstruction tasks.

The performance of brain image recovery is evaluated using two metrics: root mean squared error (RMSE) and structural similarity index Measure (SSIM)~\cite{wang2004image}. RMSE quantifies the reconstruction error by measuring the difference between the reconstructed and target brain images. A lower RMSE value indicates better reconstruction accuracy, signifying minimal deviation from the ground truth. SSIM, on the other hand, assesses the similarity between the reconstructed and target images by comparing luminance, contrast, and structural attributes. A higher SSIM score reflects greater similarity and better preservation of the original image's structural and perceptual quality.

Data normalization is applied to both the training and testing datasets to scale the input fragment data to the range [0, 1]. The model is trained using the ADAM optimizer~\cite{kingma2014adam_arxiv}, with an initial learning rate set to 0.0005. The training process runs for 1000 epochs to ensure convergence. All hidden layers are configured with a size of 32, providing sufficient capacity for feature representation.
The implementation of the model is carried out using TensorFlow 2.6 and Keras frameworks, leveraging the computational power of an A100 GPU for efficient training and testing. The hyperparameter $\alpha_{perc}$ is set to 100 to balance the specific components of the loss function, ensuring stable and effective optimization.

\begin{table}[!t]
\vspace{-.2in}
\small
\newcommand{\tabincell}[2]{\begin{tabular}{@{}#1@{}}#2\end{tabular}}
\centering
\caption{Reconstruction performance by different methods in Full-Transducer (left) and Partial-Transducer (right) datasets in terms of (SSIM, RMSE). The performance is measured by the average results of the  50 slices of 2D images.}
\vspace{-.1in}
\begin{tabular}{|l|ccc|ccc|}
\hline
%\textbf{Method} & LES Based & Downsample Based &  \\ \hline 
\textbf{Method} & \textbf{Full-Transducer} & \textbf{Partial-Transducer}& \\ \hline 
% FWI& (xxx, xxx) &(xxx, xxx)&\\  
U-Net& (0.753, 0.176) &(0.721, 0.195)&\\  
RCAN& (0.761, 0.165) &(0.743, 0.186)&\\
CycleGAN& (0.770, 0.159) &(0.749, 0.181)&\\ 
LDM& (0.811, 0.137) &(0.811, 0.146)&\\  
ViT& (0.813, 0.138) &(0.811, 0.145)&\\  
BrainPuzzle& (0.849, 0.123) &(0.833, 0.129)&\\ 
\hline
\end{tabular}
\label{fig:table1}
\vspace{-.2in}
\end{table}

\begin{figure} [!t] 
\vspace{-.1in}
\centering
\subfigure[Homo Initial SoS]{ \label{fig:a}{}
\includegraphics[width=0.35\columnwidth]{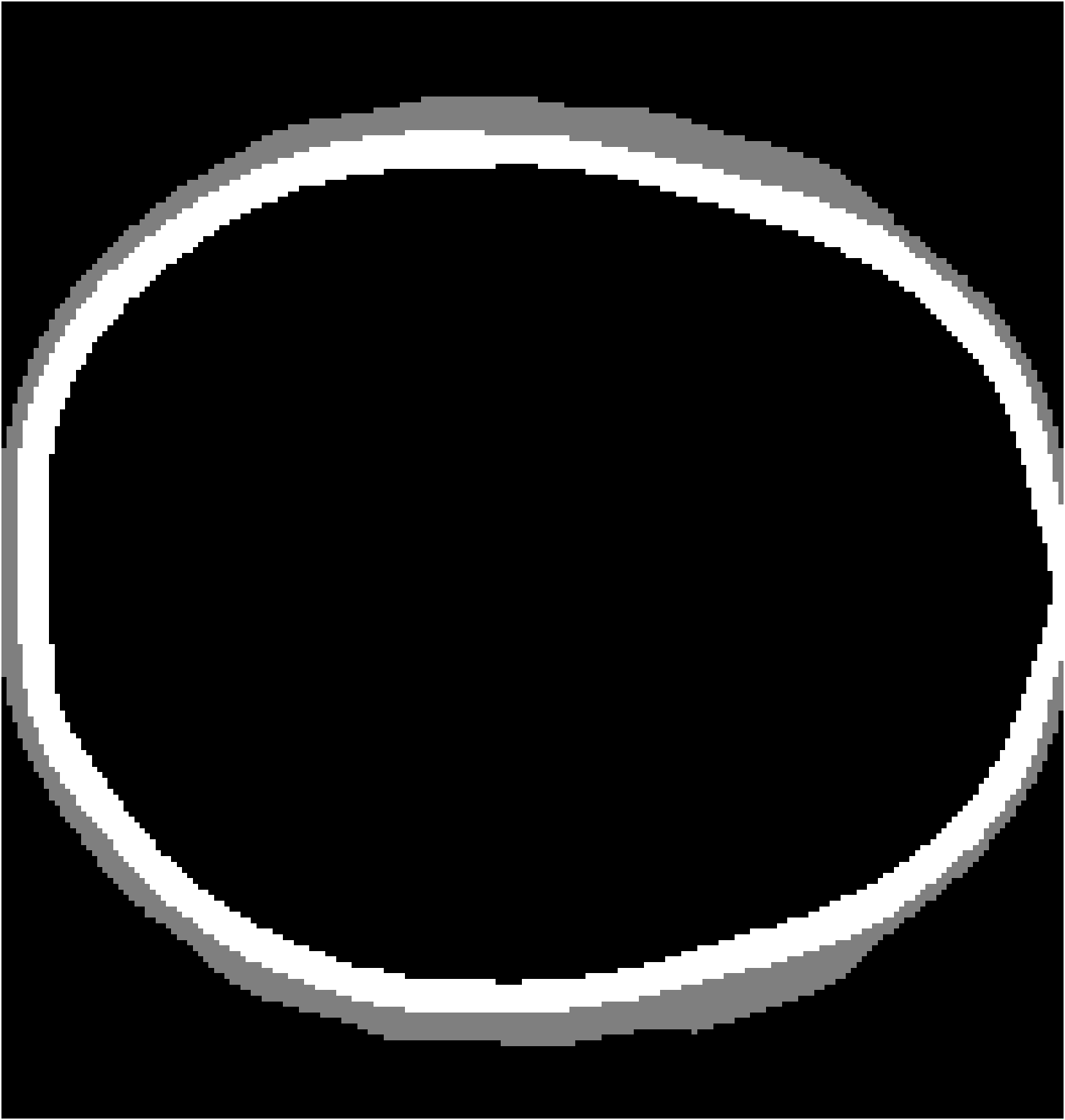}
}\hspace{1mm}
\subfigure[FWI SoS (Homo Initial)]{ \label{fig:b}{}
\includegraphics[width=0.35\columnwidth]{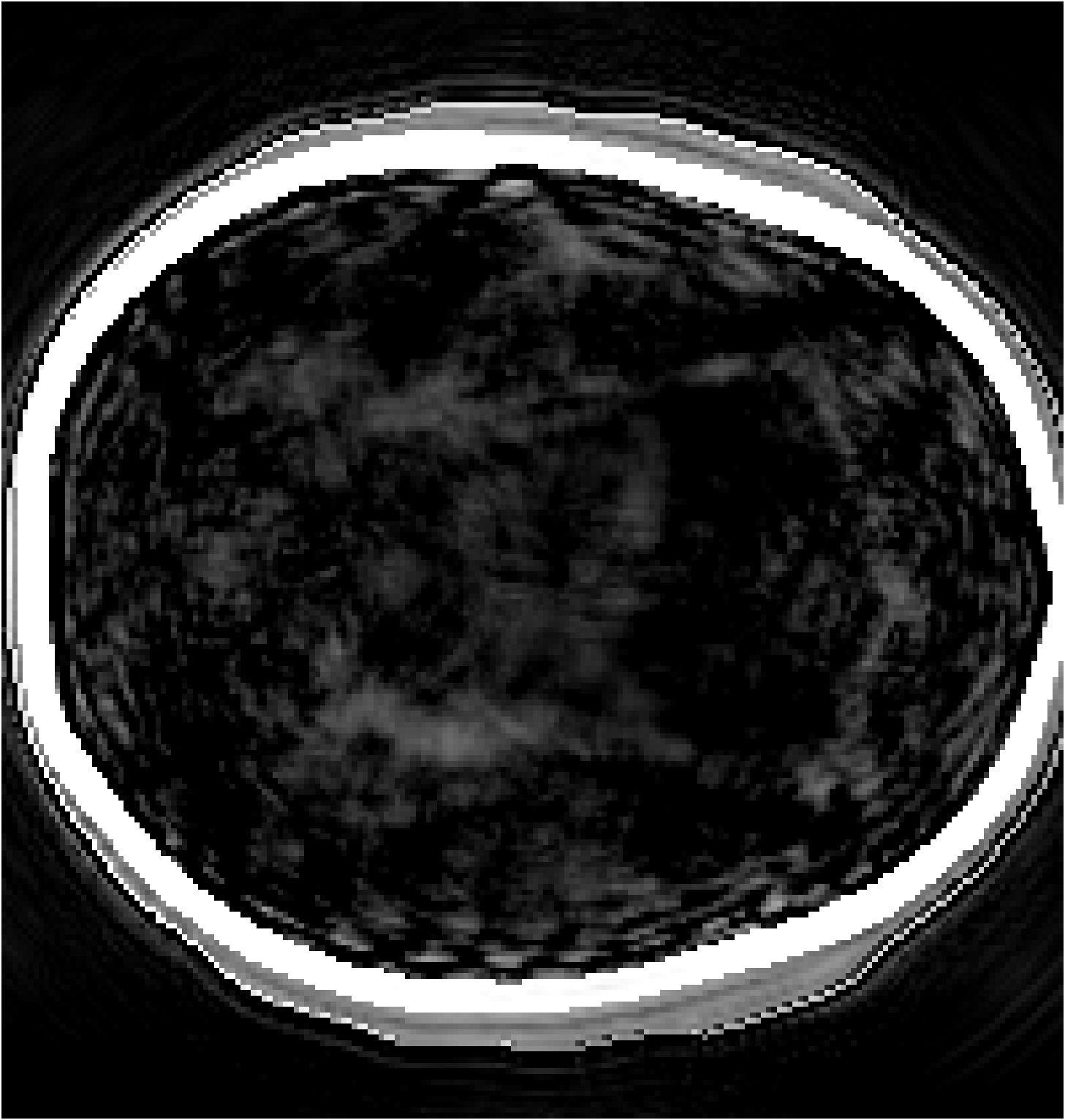}
}\vspace{-.05in}
\subfigure[Smoothed Initial SoS]{ \label{fig:a}{}
\includegraphics[width=0.35\columnwidth]{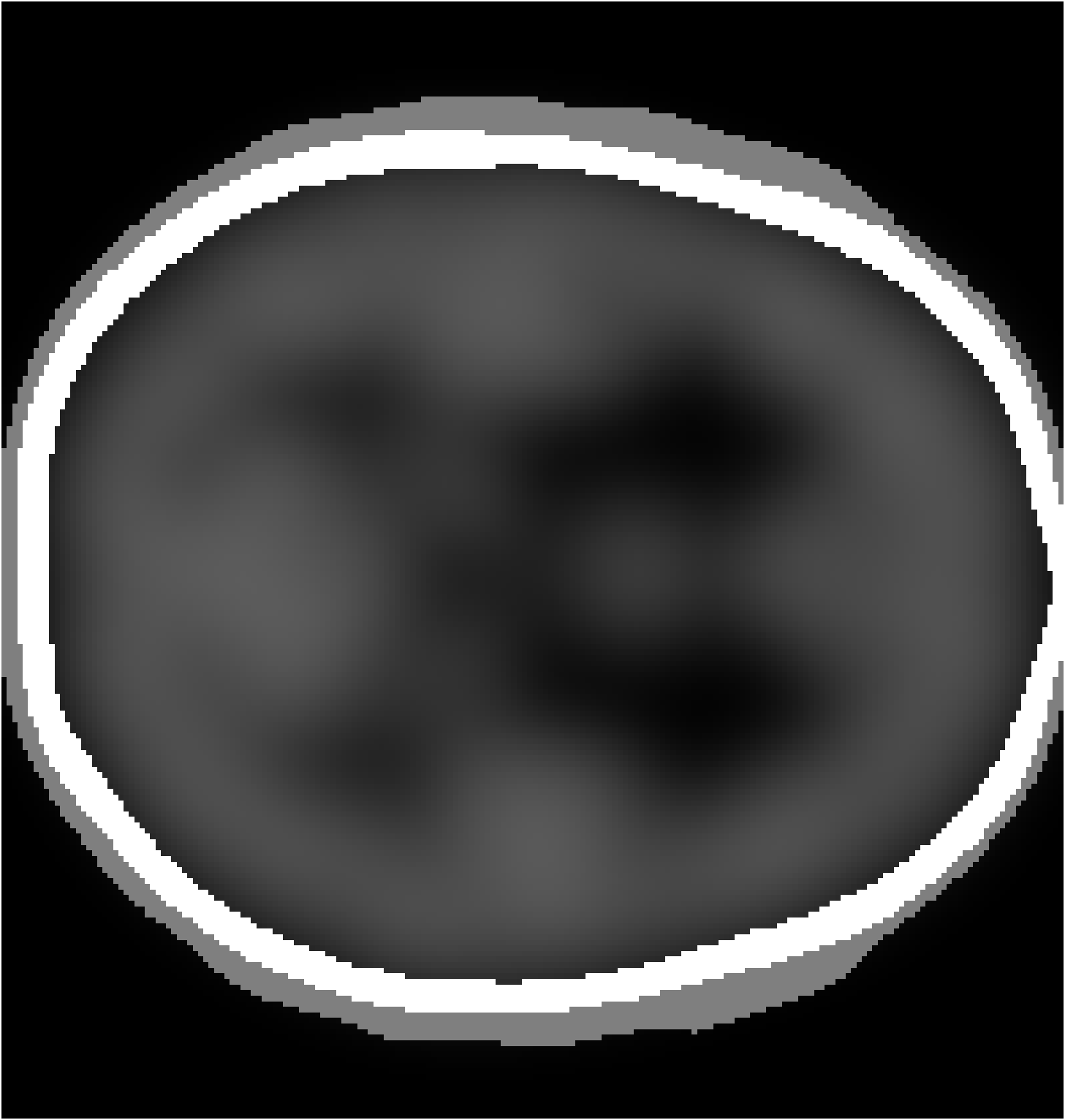}
}\hspace{1mm}
\subfigure[FWI SoS (Smoothed Initial)]{ \label{fig:b}{}
\includegraphics[width=0.35\columnwidth]{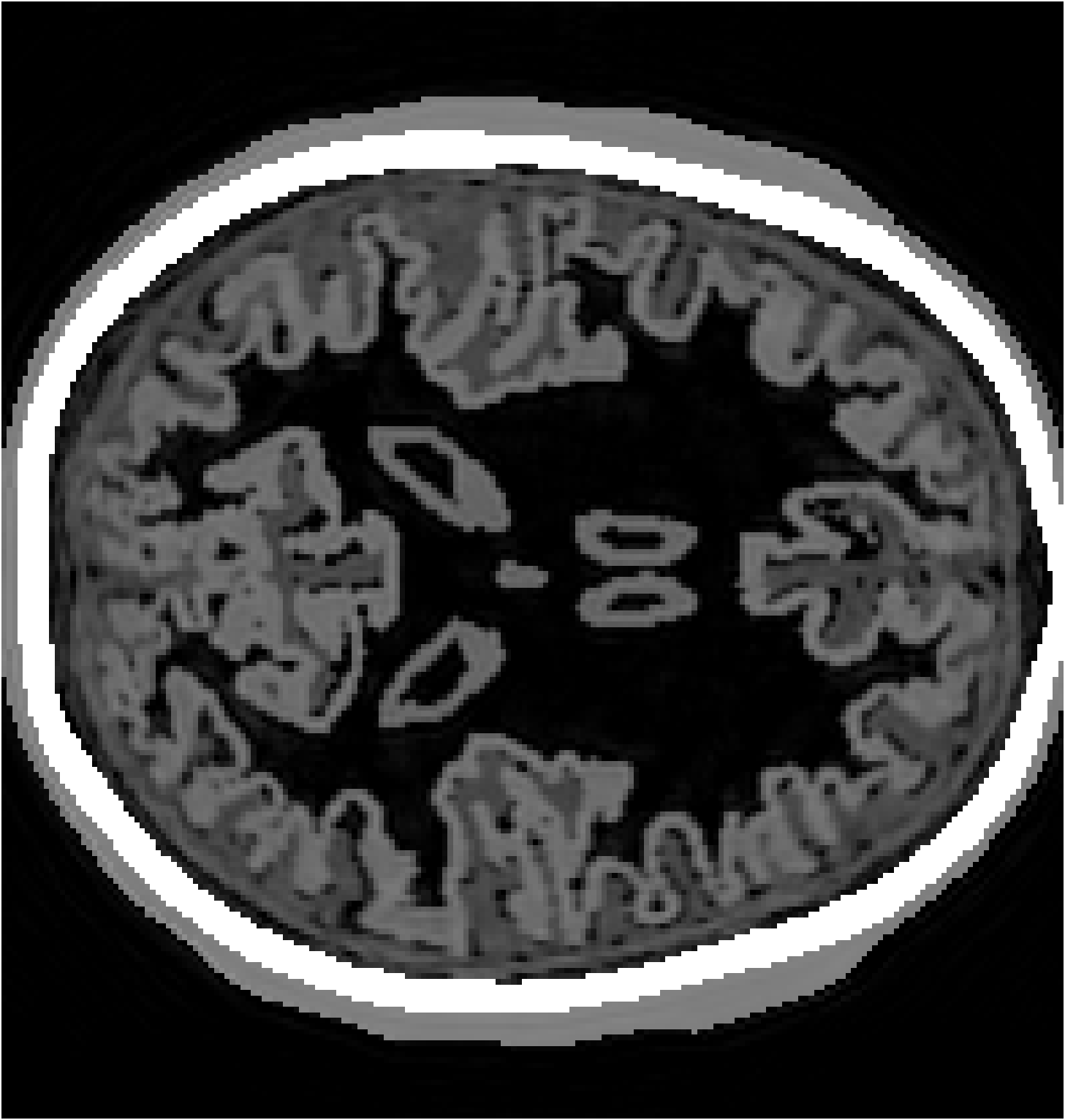}
}\vspace{-.05in}
\subfigure[BrainPuzzle SoS]{ \label{fig:c}{}
\includegraphics[width=0.35\columnwidth]{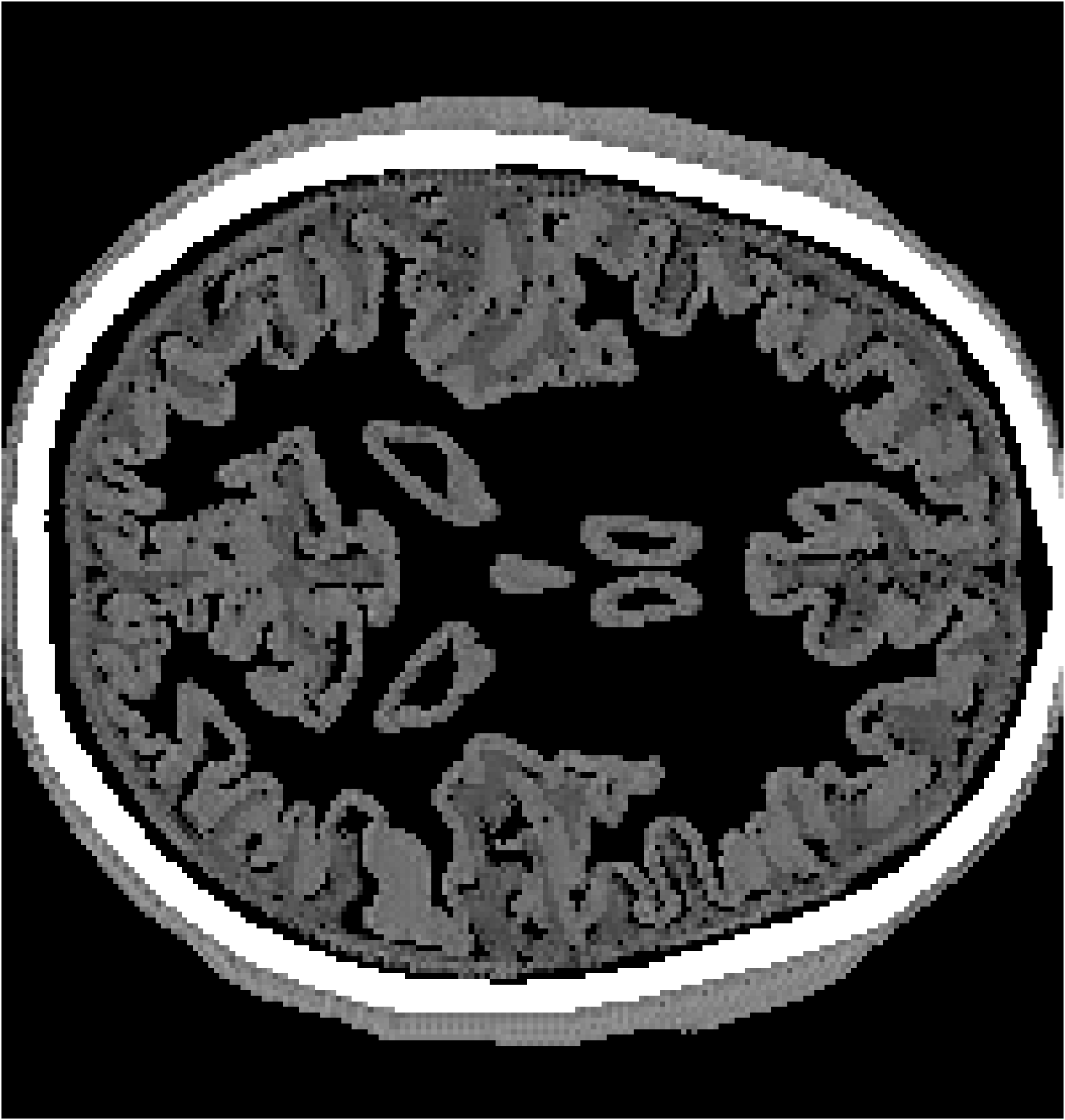}
}\hspace{1mm}
\subfigure[Ground Truth]{ \label{fig:d}{}
\includegraphics[width=0.35\columnwidth]{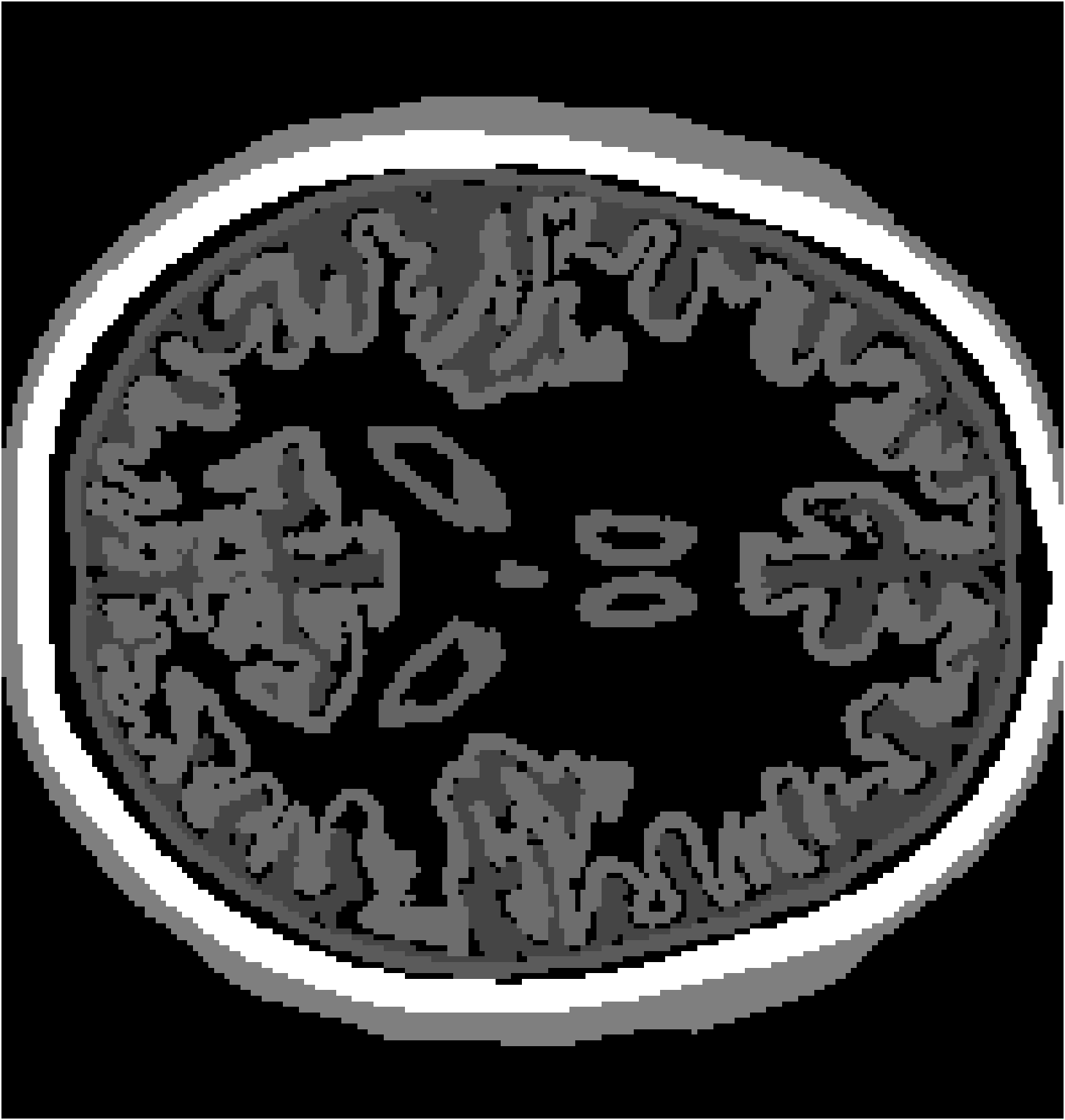}
}
% \caption{\textbf{FWI vs. BrainPuzzle.} Top left: Homogeneous Initial SoS; Top right: FWI SoS with Homogeneous Initial; Middle left: Smoothed Initial SoS; Middle right: FWI SoS with Smoothed Initial; Bottom left: BrainPuzzle SoS; Bottom right: ground truth.}
\vspace{-.1in}
\caption{\textbf{Comparison between physics-based FWI and the proposed BrainPuzzle reconstruction.} (a) Homogeneous initial SoS; (b) FWI result using homogeneous initial model; (c) Smoothed initial SoS; (d) FWI result using smoothed initial model; (e) BrainPuzzle SoS reconstruction; (f) Ground truth. BrainPuzzle yields accurate and detailed structures even without a high-quality initial model.}
\label{fig:FWI}
\vspace{-.2in}
\end{figure}

\begin{figure*} [!h]
\vspace{-.2in}
\centering
\subfigure[RCAN.]{ \label{fig:a}
\includegraphics[width=0.15\linewidth]{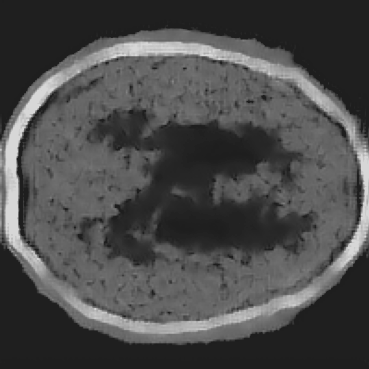}
}\hspace{2mm}
\subfigure[LDM.]{ \label{fig:b}
\includegraphics[width=0.15\linewidth]{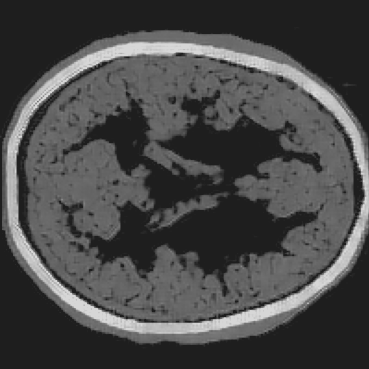}
}\hspace{2mm}
\subfigure[ViT.]{ \label{fig:c}
\includegraphics[width=0.15\linewidth]{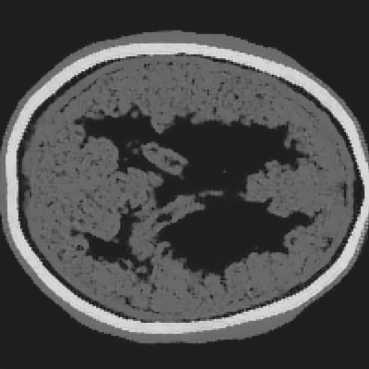}
}\hspace{2mm}
\subfigure[BrainPuzzle.]{ \label{fig:d}
\includegraphics[width=0.15\linewidth]{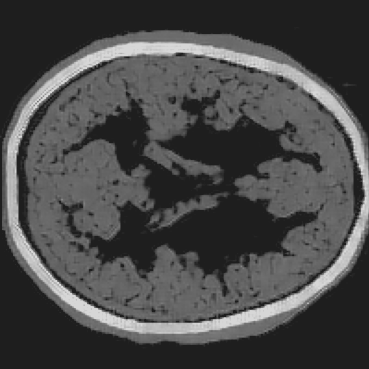}
}\hspace{2mm}
\subfigure[Ground Truth.]{ \label{fig:e}
\includegraphics[width=0.15\linewidth]{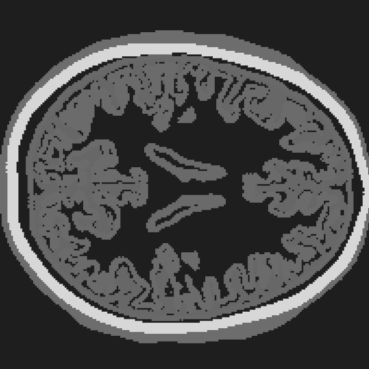}
}\vspace{-.1in}
\subfigure[RCAN.]{ \label{fig:a}
\includegraphics[width=0.15\linewidth]{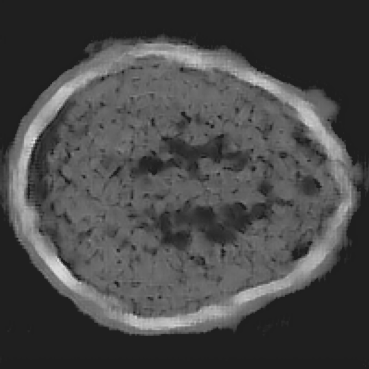}
}\hspace{2mm}
\subfigure[LDM.]{ \label{fig:b}
\includegraphics[width=0.15\linewidth]{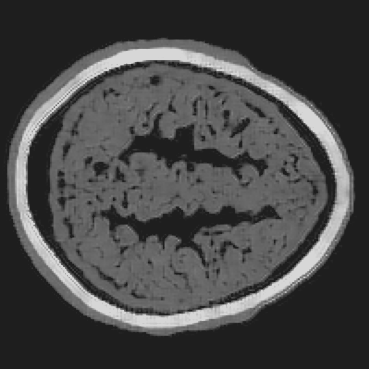}
}\hspace{2mm}
\subfigure[ViT.]{ \label{fig:c}
\includegraphics[width=0.15\linewidth]{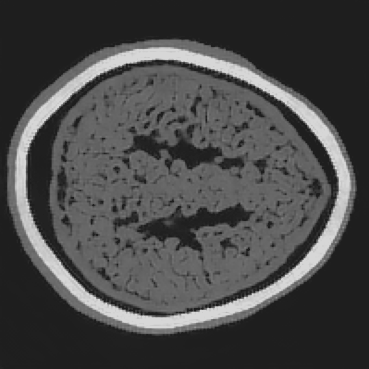}
}\hspace{2mm}
\subfigure[BrainPuzzle.]{ \label{fig:d}
\includegraphics[width=0.15\linewidth]{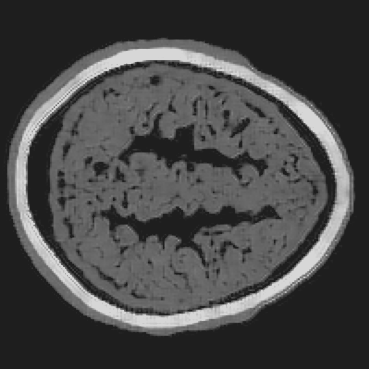}
}\hspace{2mm}
\subfigure[Ground Truth.]{ \label{fig:e}
\includegraphics[width=0.15\linewidth]{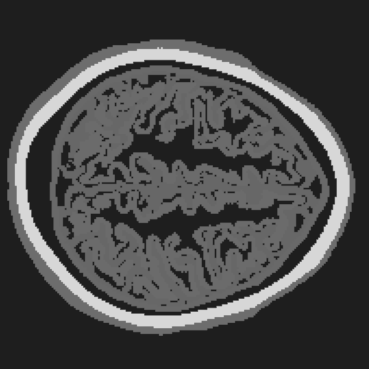}
}
\vspace{-.1in}
% \caption{Visual results of the BrainPuzzle and the three compared baselines on the partial-transducer dataset are presented, highlighting two different 2D horizontal slices.}
\caption{\textbf{Baseline comparison on partial-array data.} Two axial slices: RCAN, LDM, ViT, and BrainPuzzle versus ground truth. BrainPuzzle preserves fine structure and boundary sharpness.}
\label{fig:visual}
\vspace{-.25in}
\end{figure*}

% \subsection{Experiment analysis}
% \subsubsection{Quantitative performance and visual results}

% \textcolor{red}{Add new test to illustrate Stage 1. That will include initial velocity for TRA and the resulting TRA fragments. }

\subsection{Test 1: BrainPuzzle Quantitative Performance Analysis}

Table~\ref{fig:table1} summarizes the average performance over 50 slices of 2D brain images during the testing phase for both the full-transducer and partial-transducer datasets. The results demonstrate that the BrainPuzzle method outperforms all baseline models, achieving the highest SSIM values and the lowest RMSE values in both scenarios. Several key observations can be made: (1) %When comparing BrainPuzzle with the U-Net model and super-resolution baselines such as RCAN and CycleGAN, it is evident that these baselines struggle to effectively recover complete brain images. This limitation results in 
We observe that the U-Net model and super-resolution baselines such as RCAN and CycleGAN achieve inferior performance compared to BrainPuzzle in terms of both SSIM and RMSE. These baseline models are less capable of handling the intricate spatial relationships and texture details required for accurate reconstruction, and thus struggle to effectively recover complete brain images. 
(2) BrainPuzzle also outperforms more advanced generative models, including LDM and ViT. While LDM and ViT exhibit more complex structures and training processes, they require significantly longer training and inference times compared to BrainPuzzle. Despite their complexity, these models fail to match the accuracy and efficiency achieved by BrainPuzzle, underscoring its robustness and practical applicability. (3)~All models demonstrate significantly better performance on the full-transducer dataset compared to the partial-transducer dataset. This superiority is attributed to the clearer and more complete input brain fragments in the full-transducer scenario, which provide essential spatial information and finer-grained textures crucial for accurate reconstruction. In contrast, the partial-transducer data are not sufficient for high-quality reconstruction using limited and fragmented inputs. 

The efficacy of the BrainPuzzle method is further demonstrated by the visual results presented in Fig.~\ref{fig:visual}. These results showcase comprehensive brain images reconstructed across multiple 2D slices, highlighting the superior performance of the proposed method. A detailed analysis of these images reveals the following observations: (1) The RCAN baseline exhibits significant limitations in reconstructing the complete structural complexity of brain images. It struggles to capture the fine textures and intricate details characteristic of brain imagery, resulting in noticeably lower-quality reconstructions. (2)~BrainPuzzle achieves a significant improvement in image quality, excelling in capturing finer details such as the accuracy of structural contours and the fidelity of textural patterns. These improvements can be particularly critical for medical diagnosis and analysis, where precision and detail play a pivotal role in ensuring reliable outcomes. (3) While LDM and ViT produce reasonably good results, they fall short of BrainPuzzle in accurately reconstructing smaller details. The minor deficiencies in their reconstructions highlight BrainPuzzle’s superior ability to preserve intricate features and deliver higher-quality outputs.

Additionally, we also compare BrainPuzzle with the physics-based FWI with results illustrated in Figure~\ref{fig:FWI}. The physics-based FWI is conducted using the steepest descent method over 30 epochs with the full-transducer dataset. We use two different initial models: one with a homogeneous velocity assigned to all tissues inside the skull~(Figure~\ref{fig:FWI}(a)), and the other obtained by smoothing the ground-truth tissue distribution within the skull~(Figure~\ref{fig:FWI}(b)). The model with a homogeneous initial condition produces poor results, while the one with a smoothed initial model yields much better outcomes, indicating that the physics-based method relies heavily on the accuracy of the initial guess of SoS. In contrast, Brain Puzzle SoS achieves good results even with the partial-transducer setup as shown in Figure~\ref{fig:FWI}(e).

% \subsection{Test 2: Comparison with Physics Method: Accuracy and Computational Cost}

% \textcolor{red}{Please further elaborate more details on the comparison with physics methods. That will include all the details of the physics-based FWI method. A detail discussion on the results, including both visual and quantitative analysis (SSIM, RMSE), and others.} 

% \textcolor{red}{Also, remember to discuss on the computational cost of using physics-based and BrainPuzzle.}

% The physical method, FWI, fails to achieve satisfactory performance, as illustrated in Fig.~\ref{fig:FWI}.%, resulting in decreased reconstruction quality across all models.

\begin{table}[!t]
\vspace{-.2in}
\small
\newcommand{\tabincell}[2]{\begin{tabular}{@{}#1@{}}#2\end{tabular}}
\centering
\caption{Reconstruction performance in the ablation study is evaluated on both datasets in terms of SSIM and RMSE. The performance is measured based on the average results of 50 slices of 2D images.}
\vspace{-.1in}
\begin{tabular}{|l|ccc|ccc|}
\hline
%\textbf{Method} & LES Based & Downsample Based &  \\ \hline 
\textbf{Method} & \textbf{Full-Transducer} & \textbf{Partial-Transducer}& \\ \hline 
$\text{BrainPuzzle}_\text{T}$& (0.761, 0.165) &(0.743, 0.186)&\\  
$\text{BrainPuzzle}_\text{G}$& (0.835, 0.129) &(0.821, 0.135)&\\  
BrainPuzzle& (0.849, 0.123) &(0.833, 0.129)&\\ 
\hline
\end{tabular}
\label{fig:table2}
\vspace{-.2in}
\end{table}

\subsection{Test 2: Ablation Study}

% \subsubsection{Ablation study}
Table~\ref{fig:table2} presents the results of the ablation study conducted on both the full-transducer and partial-transducer datasets. A comparative analysis among the $\text{BrainPuzzle}_\text{T}$, $\text{BrainPuzzle}_\text{G}$, and the full BrainPuzzle model highlights the impact of each proposed component: transformer-based super-resolution framework, GAU, and the automatic attention mechanism. The results demonstrate that incorporating each of these components significantly improves reconstruction performance. Notably, the inclusion of the transformer-based super-resolution framework and GAU contributes the most substantial enhancements, as evidenced by the improved SSIM and RMSE metrics. These findings underscore the critical role of these components in capturing intricate spatial relationships and refining the high-resolution details necessary for accurate brain image reconstruction.

\begin{figure}
%\vspace{-.1in}
\centering
\includegraphics[width=0.9\columnwidth]{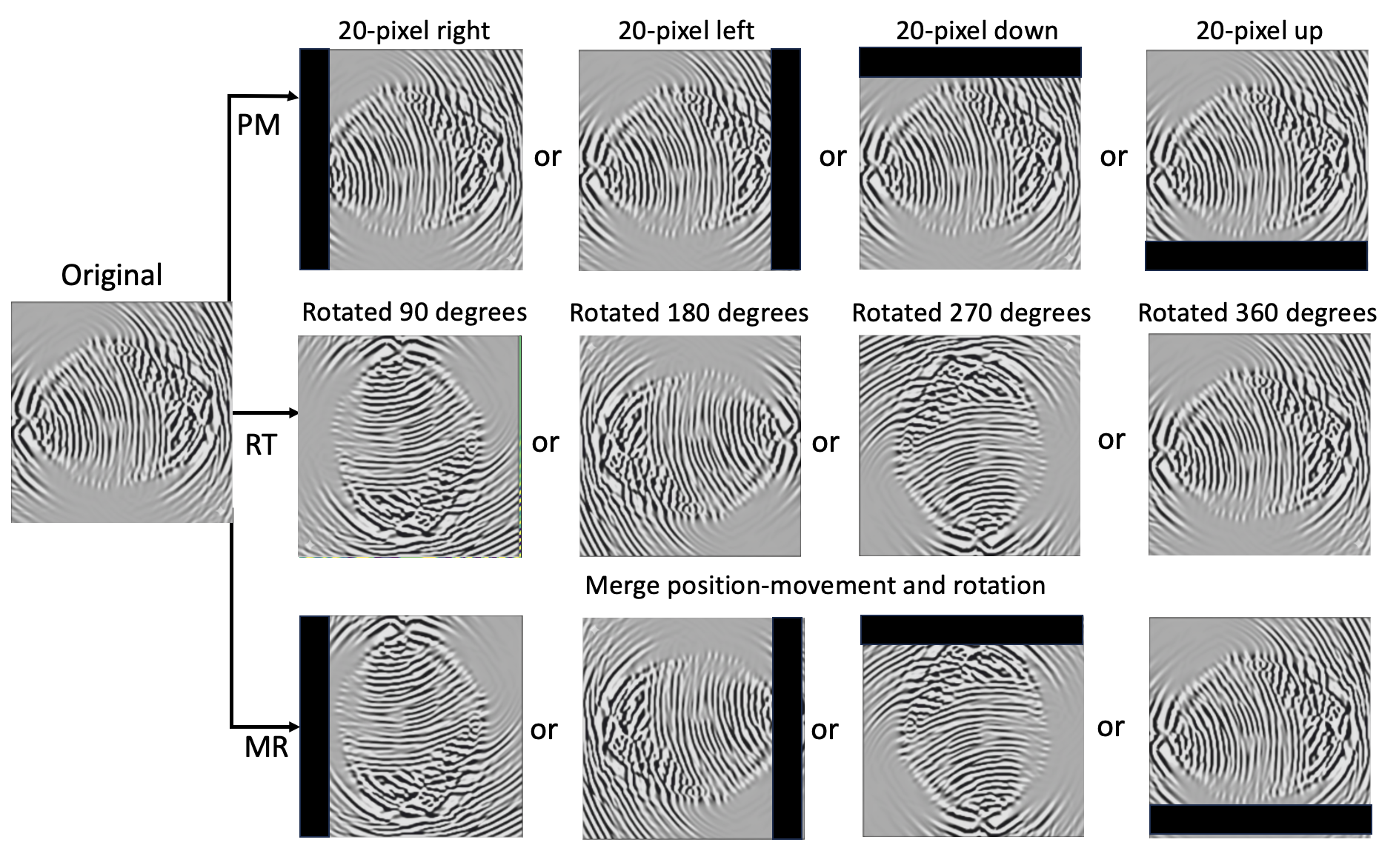}
% \caption{Pre-processing of fragments through position movement (PM), rotation (RT), and moving-rotation (MR).~(\textcolor{red}{this figure needs revision.})}
\vspace{-.2in}
\caption{\textbf{Fragment perturbations.} Position-moved (PM), rotated (RT), and moved+rotated (MR) variants used to assess robustness, compared to the original fragments.}
\label{fig:fragments}
\vspace{-.2in}
\end{figure}

% \subsubsection{Generalizability}
\subsection{Test 3: Generalizability Study}

In this study, we develop a comprehensive experimental framework to rigorously evaluate the generalization and practical applicability of the BrainPuzzle method. The primary objective is to assess the model's ability to reconstruct complete brain images from a diverse set of brain fragments, simulating real-world challenges encountered in practical scenarios. To achieve this, we generate various types of brain fragments from ultrasound data, specifically designed to replicate common challenges. These fragment types include original fragments (Original), position-moving fragments (PM), rotated fragments (RT), and moving-rotated fragments (MR). As shown in the Figure~\ref{fig:fragments}, original fragments are generated directly from ultrasound data and typically exhibit overlap. They represent the standard case without any modifications. PM fragments introduce complexity by simulating scenarios where the object's position has randomly shifted (here, we shift 20 pixels) from different borders for each fragment, creating positional inconsistencies. RT fragments are randomly rotated by angles of $0^\circ$, $90^\circ$, $180^\circ$, or $270^\circ$, reflecting rotational variability in real-world data. MR fragments combine both positional movement and rotation, representing the most challenging case, as they involve simultaneous translation and orientation changes. The PM, RT, and MR fragments are all derived as modifications of the original fragments. Despite these variations, the datasets for all fragment types share the same foundation: 150 slices of 2D speed-of-sound images, with each slice containing 50 brain fragments representing different scenarios.
For the experiments, we maintain the same setup as in previous tests. From each dataset, 100 original samples are randomly selected for model training, while the remaining 50 samples are used for evaluation across different fragment cases. This experimental design ensures consistency while allowing us to systematically assess the model's performance under varying levels of complexity.

\begin{figure} [!t] 
\vspace{-.15in}
\centering
\subfigure[]{ \label{fig:b}{}
\includegraphics[width=0.4\columnwidth]{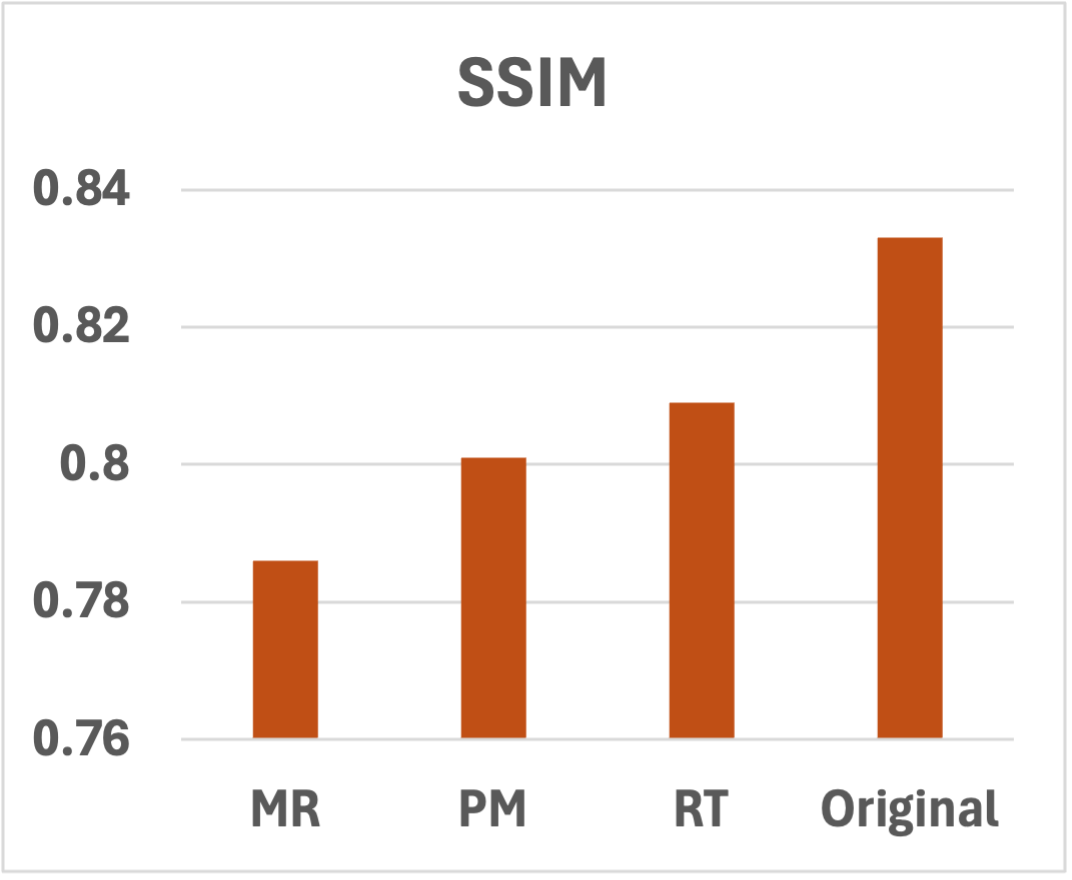}
}\hspace{.1in}
\subfigure[]{ \label{fig:b}{}
\includegraphics[width=0.4\columnwidth]{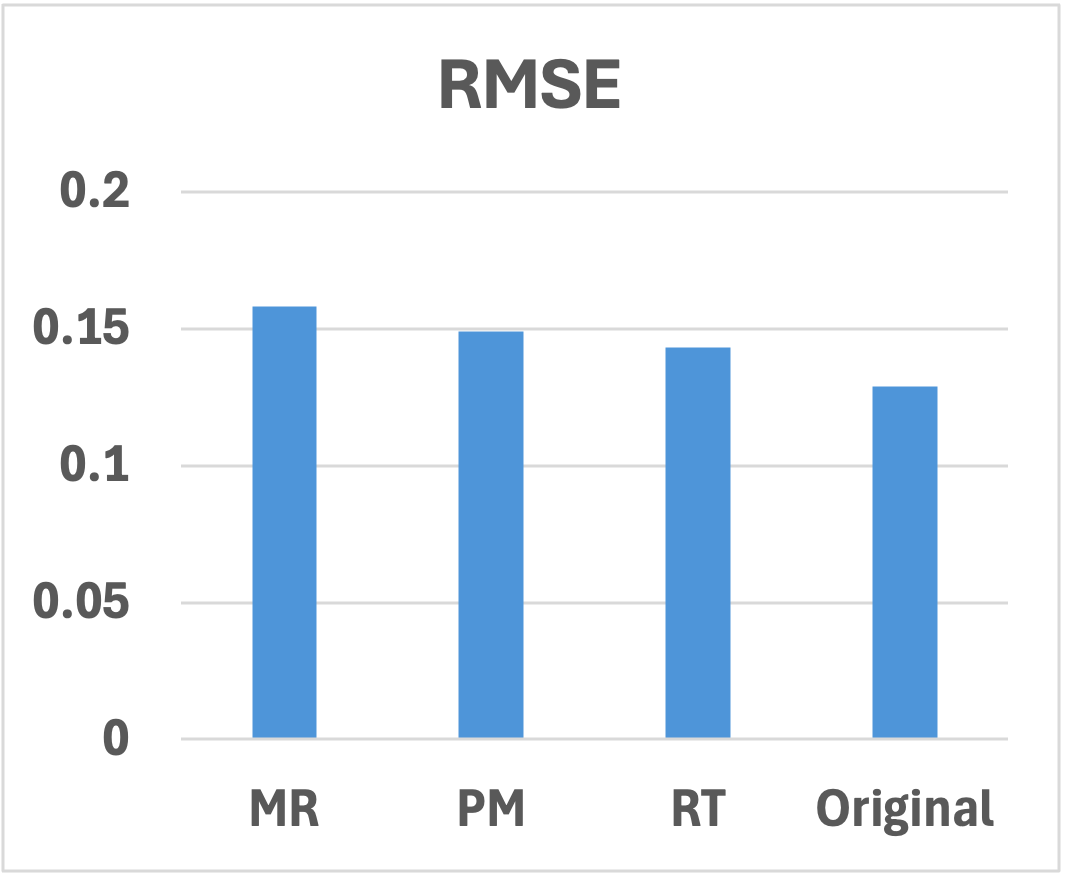}
}%\vspace{-.15in}
% \caption{Reconstruction performance (measured by RMSE and SSIM) of the BrainPuzzle method on different cases of fragments input, including original fragments (Original), rotated (RT), position-moved fragments (PM), and position-moved and rotated fragments (MR). }
\vspace{-.1in}
\caption{\textbf{Robustness to fragment perturbations.} SSIM (left) and RMSE (right) for Original, RT, PM, and MR inputs. Performance degrades gracefully as spatial perturbations increase.}
\label{fig:fragmentsResults}
\vspace{-.2in}
\end{figure}

\begin{figure*}[t]
\vspace{-.2in}
\centering
\subfigure[MR.]{ \label{fig:a}
\includegraphics[width=0.17\linewidth]{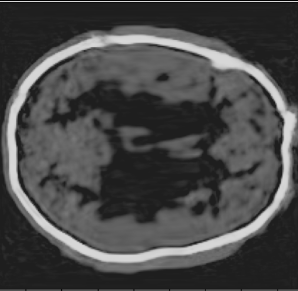}
}%\hspace{1mm}
\subfigure[PM.]{ \label{fig:b}
\includegraphics[width=0.17\linewidth]{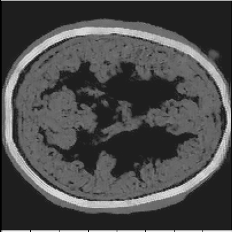}
}%\hspace{2mm}
\subfigure[RT.]{ \label{fig:c}
\includegraphics[width=0.17\linewidth]{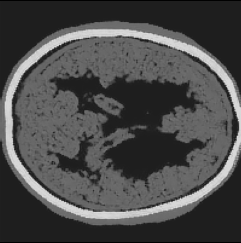}
}%\hspace{2mm}
\subfigure[Original.]{ \label{fig:d}
\includegraphics[width=0.17\linewidth]{Image1/Brain_i1.png}
}%hspace{2mm}
\subfigure[Ground Truth.]{ \label{fig:e}
\includegraphics[width=0.17\linewidth]{Image1/GT1.png}
}
\vspace{-.1in}
% \caption{Visual results of BrainPuzzle on the partial-transducer dataset are shown for different cases of fragment inputs, including original fragments (Original), rotated (RT), position-moved fragments (PM), and position-moved and rotated fragments (MR).}
\caption{\textbf{Qualitative robustness.} Reconstructions under MR, PM, RT, and Original fragment inputs compared to ground truth on partial-array data.}
\label{fig:visual1}
\vspace{-.2in}
\end{figure*}
Figures~\ref{fig:fragmentsResults} present quantitative comparisons across different types of brain image fragments: original fragments, PM fragments, RT fragments, and MR fragments. The experimental results demonstrate the superior performance of the BrainPuzzle method, particularly in terms of RMSE and SSIM values. Among the cases, the original fragment scenario achieves the best performance, reflecting an ideal condition where most of the necessary spatial and contextual information for complete image reconstruction is retained in the input data.
In contrast, the performance in the PM and RT cases is slightly diminished. This reduction is attributed to the distortion of critical spatial relationships and the loss of some contextual cues caused by positional movement in the PM fragments and rotational variability in the RT fragments. Although BrainPuzzle effectively mitigates these challenges to a significant extent, the degradation in performance indicates that these distortions affect the model’s ability to fully reconstruct the original image. The MR case, however, records the lowest performance, presenting the most complex reconstruction challenge. This case combines positional movement with rotational transformations, leading to a substantial alteration of the original spatial context. These compounded changes disrupt the continuity of textures and patterns in the brain fragments, making it significantly more difficult for the model to align and reconstruct the fragments into a coherent image. The extensive spatial disorientation and loss of critical information in the MR fragments underscore the limitations of the BrainPuzzle method when faced with severe disruptions in spatial integrity. 

Similar observations can be observed in Figure~\ref{fig:visual1}, which provides a visual comparison of the reconstructed images across the different fragment scenarios. The images highlight how the original fragments yield the most accurate reconstructions, while the MR fragments result in less precise outputs, further emphasizing the challenges posed by the most complex case. These results collectively demonstrate the robustness of BrainPuzzle under varying conditions and identify areas for further improvement in handling extreme distortions.

\begin{figure} [!t] 
% \vspace{-.1in}
\centering
\subfigure[]{ \label{fig:b}{}
\includegraphics[width=0.4\columnwidth]{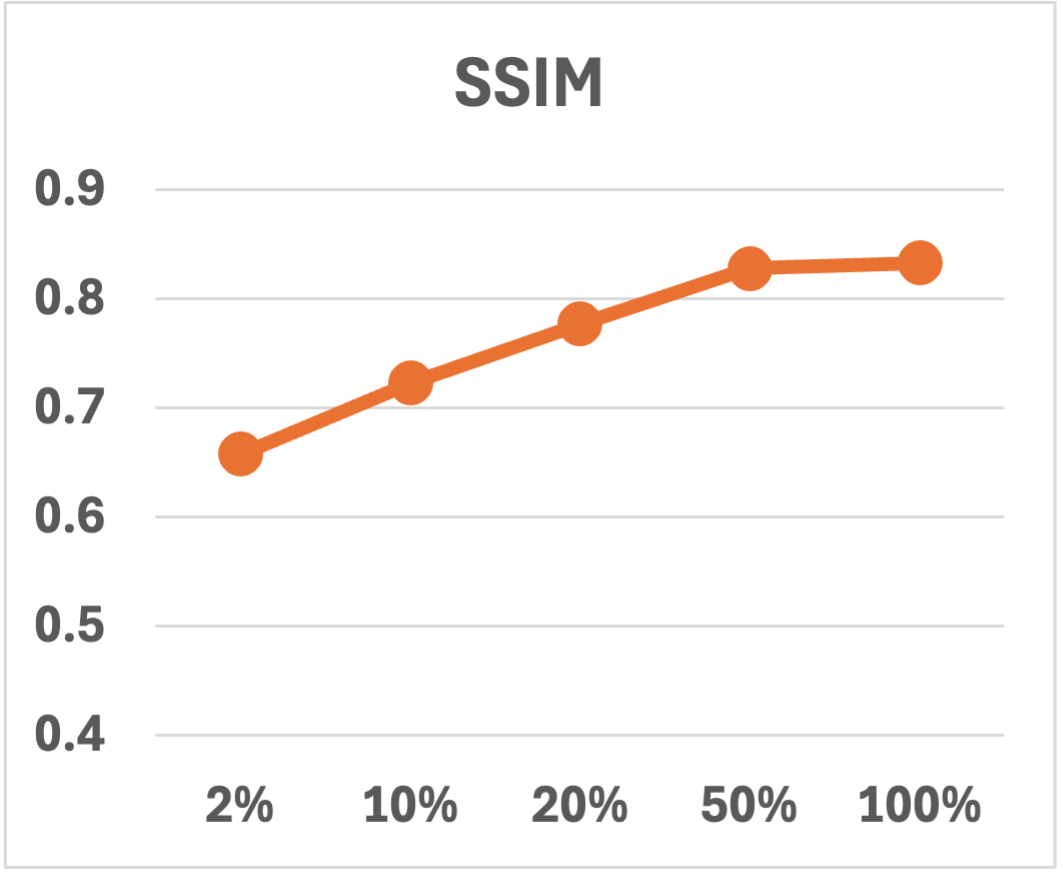}
}\hspace{.1in}
\subfigure[]{ \label{fig:b}{}
\includegraphics[width=0.4\columnwidth]{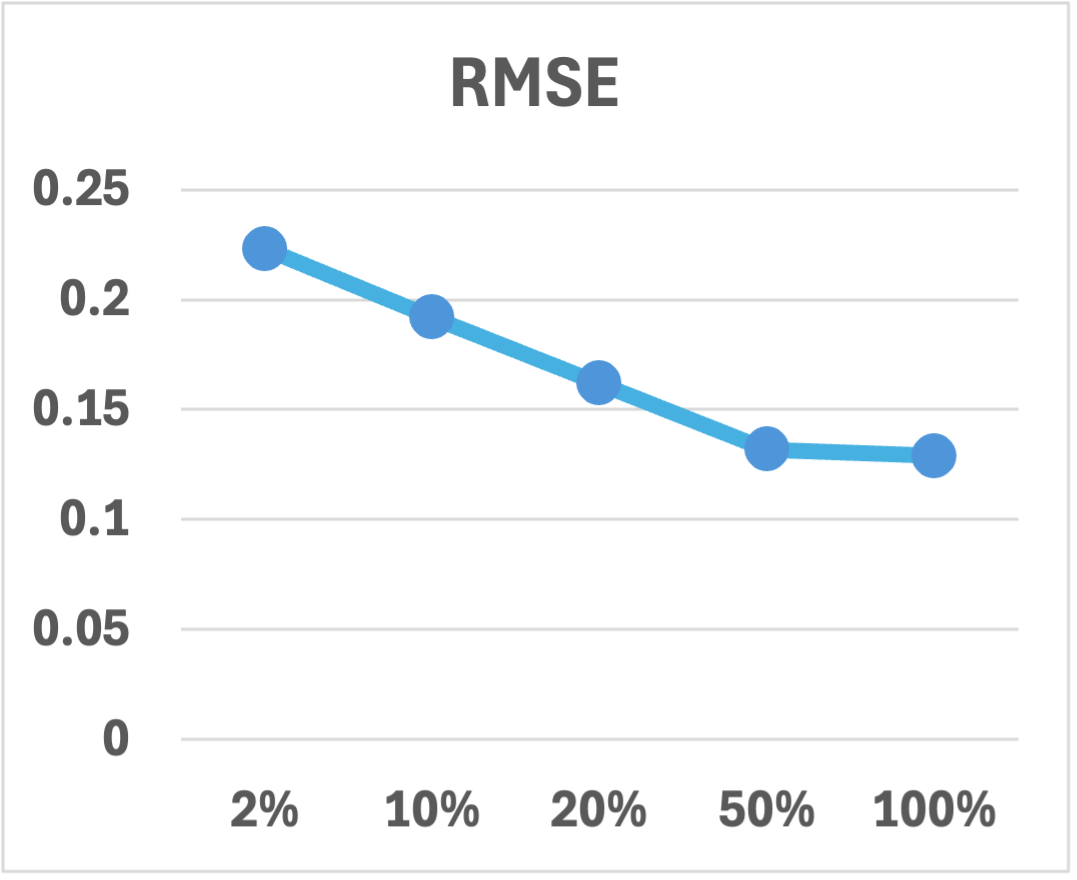}
}%\vspace{-.15in}
% \caption{Reconstruction performance of the BrainPuzzle method (measured by RMSE and SSIM) with different percentages of fragment inputs.}
\vspace{-.1in}
\caption{\textbf{Effect of fragment sparsity.} SSIM (left) and RMSE (right) versus percentage of available fragments. BrainPuzzle remains accurate down to moderate sparsity, with expected decline at extreme sparsity.}
\label{fig:sparse}
\vspace{-.2in}
\end{figure}

Figure~\ref{fig:sparse} illustrates the performance of the BrainPuzzle method with varying percentages of fragment inputs. The results clearly show that the model achieves significant performance improvements even with a relatively small percentage of fragments (e.g., 50\%), demonstrating its robustness in handling partial information. However, as the percentage of input fragments decreases to 10\% or 2\%, there is a noticeable decline in SSIM and a corresponding increase in RMSE, highlighting the model's limitations in scenarios with sparse input data. This behavior underscores the critical importance of providing sufficient fragment coverage to ensure high-quality reconstructions. At the same time, the results showcase the model’s resilience and ability to adapt to suboptimal conditions, making it effective even with limited data. These findings emphasize the versatility of BrainPuzzle across different fragment availability scenarios and its potential applicability in real-world settings where incomplete data is a frequent challenge.

% In addition, Table~\ref{fig:table3} presents the performance of BrainPuzzle under various levels of Gaussian noise injected into the fragments. As noise intensity increases from 
% $\mathcal{N}(0,0.00)$ to $\mathcal{N}(0,0.20)$, both SSIM and RMSE metrics indicate a gradual degradation in reconstruction quality for both full-transducer and partial-transducer settings. Notably, the full-transducer configuration consistently outperforms the partial-transducer across all noise levels, suggesting that the presence of multiple transducers provides greater robustness against noise. Nevertheless, BrainPuzzle maintains a reasonable performance even at higher noise levels (e.g., SSIM of 0.681 with RMSE of 0.215 under $\mathcal{N}(0,0.20)$), demonstrating its resilience in noisy environments. These results confirm the BrainPuzzle's capability to generalize well under imperfect and noisy inputs, which is critical for real-world applications where measurement errors and noise are often unavoidable.

\subsection{Test 4: Noise Robustness Study}

Table~\ref{fig:table3} illustrates the robustness of BrainPuzzle by evaluating its reconstruction performance under varying levels of Gaussian noise added to input fragments in a practical, partial-transducer configuration. As the noise intensity increases from $\mathcal{N}(0,0.00)$ to $\mathcal{N}(0,0.20)$, both SSIM and RMSE metrics indicate a gradual decline in reconstruction accuracy. Despite this degradation, BrainPuzzle demonstrates consistent resilience, maintaining reasonable performance even at higher noise intensities (e.g., SSIM of 0.653 with RMSE of 0.224 under $\mathcal{N}(0,0.20)$). This robust performance in noisy scenarios is crucial for practical clinical applications where measurement inaccuracies and noise are inevitable. For comparison, results from the full-transducer configuration are also provided, which consistently outperform the partial-transducer setup across all noise levels. However, as previously noted, deploying the full-transducer arrangement is impractical due to its prohibitive cost, complexity, and computational demands, reinforcing the necessity and relevance of evaluating and optimizing the partial-transducer approach.

\begin{table}[!t]
% \vspace{-.1in}
\small
\newcommand{\tabincell}[2]{\begin{tabular}{@{}#1@{}}#2\end{tabular}}
\centering
\caption{BrainPuzzle's performance (measured by SSIM and RMSE) by containing different levels of random Gaussian noise $\mathcal{N}$ in the fragments.}
\vspace{-.1in}
\begin{tabular}{|l|ccc|ccc|}
\hline
%\textbf{Method} & LES Based & Downsample Based &  \\ \hline 
\centering{\textbf{Noise Level}} & \textbf{Full-Transducer} & \textbf{Partial-Transducer}&  \\ \hline 
$\mathcal{N}(0,0.00)$& (0.849, 0.123) &(0.833, 0.129)&\\  
$\mathcal{N}(0,0.01)$& (0.845, 0.127) &(0.828, 0.133)&\\  
$\mathcal{N}(0,0.02)$& (0.838, 0.129) &(0.823, 0.133)&\\
$\mathcal{N}(0,0.05)$& (0.821, 0.135) &(0.817, 0.141)&\\ 
$\mathcal{N}(0,0.10)$& (0.807, 0.146) &(0.803, 0.152)&\\ 
$\mathcal{N}(0,0.15)$& (0.765, 0.164) &(0.743, 0.171)&\\  
$\mathcal{N}(0,0.20)$& (0.681, 0.215) &(0.653, 0.224)&\\ 
\hline
\end{tabular}
\label{fig:table3}
\vspace{-.2in}
\end{table}

\section{Conclusion and Future Work}
We developed BrainPuzzle, a hybrid two-stage framework for quantitative transcranial ultrasound imaging. In the first stage, a physics-driven TRA module converts low-SNR, limited-aperture channel data into migration fragments with reliable kinematics. The second stage fuses these fragments using a transformer encoder–decoder equipped with super-resolution and a graph-based attention unit (GAU) to reconstruct a quantitative speed-of-sound (SoS) map. Evaluations on a numerical brain phantom under both full-aperture and partial-array acquisitions show that BrainPuzzle outperforms physics-based FWI and strong learned baselines (U-Net, RCAN, ViT, LDM) , producing sharper interfaces and demonstrating robustness to fragment sparsity, view perturbations, and noise. Ablation studies confirm that the transformer-based super-resolution and GAU components are essential, enabling accurate SoS reconstruction with fewer transducers and reduced hardware complexity.

Looking ahead, we plan to extend BrainPuzzle along three main directions. First, we will develop a 3D volumetric version capable of integrating multi-plane fragment data for anatomically consistent brain reconstruction. Second, we aim to adapt the model to clinical ultrasound data through domain adaptation and self-supervised learning to bridge the gap between synthetic and real-world conditions. Finally, we will explore multi-modal ultrasound imaging by incorporating additional acoustic biomarkers—such as tissue density, attenuation, and elasticity—to provide more comprehensive and clinically meaningful brain assessments.

% In this paper, we present BrainPuzzle, a novel data-driven method designed to reconstruct complete brain images from limited ultrasound fragments. The BrainPuzzle framework comprises two core components: the transformer-based super-resolution framework and the graph-based attention unit (GAU). Together, these components enable the model to effectively process and integrate fragmented image data. Additionally, we describe the process of generating ultrasound fragments using the time reversal acoustics (TRA) method, highlighting its role in simulating realistic input data for the reconstruction task. Experimental results on two distinct datasets validate the capabilities of the BrainPuzzle model and demonstrate the effectiveness of its individual components. Furthermore, we extend BrainPuzzle from 2D simulation experiments to 3D practical scenarios, showcasing its adaptability and potential for real-world applications. This transition underscores the model's scalability and robustness when applied to more complex imaging cases. The adoption of image puzzle reconstruction and super-resolution techniques in this study introduces a promising new approach to brain imaging recovery research. By bridging advanced transformer architectures with graph-based learning, BrainPuzzle establishes the foundation for future innovations in the field of medical imaging.

\section{Acknowledgements}
Y. Lin acknowledges support from the School of Data Science and Society at the University of North Carolina through a faculty start-up grant, and from the U.S. National Science Foundation Division of Mathematical Sciences. We also gratefully acknowledge a constructive discussion with Dr.~Felix J. Herrmann, whose insights have meaningfully informed and helped reshape this work.

\bibliographystyle{IEEEtran}
\bibliography{IEEEabrv-SC, sample-base1, sample-base}

% Generated by IEEEtran.bst, version: 1.12 (2007/01/11)
\begin{thebibliography}{10}
\providecommand{\url}[1]{#1}
\csname url@samestyle\endcsname
\providecommand{\newblock}{\relax}
\providecommand{\bibinfo}[2]{#2}
\providecommand{\BIBentrySTDinterwordspacing}{\spaceskip=0pt\relax}
\providecommand{\BIBentryALTinterwordstretchfactor}{4}
\providecommand{\BIBentryALTinterwordspacing}{\spaceskip=\fontdimen2\font plus
\BIBentryALTinterwordstretchfactor\fontdimen3\font minus \fontdimen4\font\relax}
\providecommand{\BIBforeignlanguage}[2]{{%
\expandafter\ifx\csname l@#1\endcsname\relax
\typeout{** WARNING: IEEEtran.bst: No hyphenation pattern has been}%
\typeout{** loaded for the language `#1'. Using the pattern for}%
\typeout{** the default language instead.}%
\else
\language=\csname l@#1\endcsname
\fi
#2}}
\providecommand{\BIBdecl}{\relax}
\BIBdecl

\bibitem{Neuroscience-2024-Triana}
A.~Triana, J.~Saramaki, E.~Glerean, and M.~Nicholas~Hayward, ``Neuroscience meets behavior: A systematic literature review on magnetic resonance imaging of the brain combined with real-world digital phenotyping,'' \emph{Human Brain Mapping}, vol.~45, no.~4, p. e26620, 2024.

\bibitem{Twenty-2012-Bandettini}
P.~Bandettini, ``Twenty years of functional mri: The science and the stories,'' \emph{NeuroImage}, vol.~62, no.~2, pp. 575--588, 2012.

\bibitem{Impact-2024-Choi}
S.~Y. Choi, J.~H. Kim, H.~S. Chung, S.~Lim, E.~H. Kim, and A.~Choi, ``Impact of a deep learning-based brain ct interpretation algorithm on clinical decision-making for intracranial hemorrhage in the emergency department,'' \emph{Scientific Reports}, vol.~14, no. 22292, 2024.

\bibitem{Computed-2022-Hillal}
A.~Hillal, T.~Ullberg, B.~Ramgren, and J.~Wasselius, ``Computed tomography in acute intracerebral hemorrhage: neuroimaging predictors of hematoma expansion and outcome,'' \emph{Insights into Imaging}, vol.~13, no. 180, 2022.

\bibitem{Transcranial-2013-Naqvi}
J.~Naqvi, K.~H. Yap, G.~Ahmad, and J.~Ghosh, ``Transcranial doppler ultrasound: A review of the physical principles and major applications in critical care,'' \emph{International Journal of Vascular Medicine}, 2013.

\bibitem{wells2006ultrasound}
P.~N. Wells, ``Ultrasound imaging,'' \emph{Physics in medicine \& biology}, vol.~51, no.~13, p. R83, 2006.

\bibitem{Physics-2025-Lin}
\BIBentryALTinterwordspacing
Y.~Lin, S.~Feng, J.~Theiler, Y.~Chen, U.~Villa, J.~Rao, J.~Greenhall, C.~Pantea, M.~Anastasio, and B.~Wohlberg, ``Physics and deep learning in computational wave imaging,'' 2024. [Online]. Available: \url{https://arxiv.org/abs/2410.08329}
\BIBentrySTDinterwordspacing

\bibitem{virieux2017introduction}
J.~Virieux, A.~Asnaashari, R.~Brossier, L.~M{\'e}tivier, A.~Ribodetti, and W.~Zhou, ``An introduction to full waveform inversion,'' in \emph{Encyclopedia of exploration geophysics}.\hskip 1em plus 0.5em minus 0.4em\relax Society of Exploration Geophysicists, 2017, pp. R1--1.

\bibitem{Dual-2023-Caradoc}
T.~Caradoc, C.~Cueto, J.~Cudeiro, O.~Bates, O.~Agudo, G.~Strong, L.~Guasch, M.~Warner, and M.-X. Tang, ``Dual-probe transcranial full-waveform inversion: A brain phantom feasibility study,'' \emph{Ultrasound in Medicine and Biology}, vol.~49, no.~10, pp. 2302--2315, 2023.

\bibitem{guasch2020full}
L.~Guasch, O.~Calder{\'o}n~Agudo, M.-X. Tang, P.~Nachev, and M.~Warner, ``Full-waveform inversion imaging of the human brain,'' \emph{NPJ digital medicine}, vol.~3, no.~1, p.~28, 2020.

\bibitem{virieux2009overview}
J.~Virieux and S.~Operto, ``An overview of full-waveform inversion in exploration geophysics,'' \emph{Geophysics}, vol.~74, no.~6, pp. WCC1--WCC26, 2009.

\bibitem{baysal1983reverse}
E.~Baysal, D.~D. Kosloff, and J.~W. Sherwood, ``Reverse time migration,'' \emph{Geophysics}, vol.~48, no.~11, pp. 1514--1524, 1983.

\bibitem{wang2019full}
P.~Wang, Z.~Zhang, J.~Mei, F.~Lin, and R.~Huang, ``Full-waveform inversion for salt: A coming of age,'' \emph{The Leading Edge}, vol.~38, no.~3, pp. 204--213, 2019.

\bibitem{thiel2019comparison}
N.~Thiel, T.~Hertweck, and T.~Bohlen, ``Comparison of acoustic and elastic full-waveform inversion of 2d towed-streamer data in the presence of salt,'' \emph{Geophysical prospecting}, vol.~67, no.~2, pp. 349--361, 2019.

\bibitem{privitera2016full}
A.~Privitera, A.~Ratcliffe, and N.~Kotova, ``A full-waveform inversion case study from offshore gabon,'' in \emph{78th EAGE Conference and Exhibition 2016}, vol. 2016, no.~1.\hskip 1em plus 0.5em minus 0.4em\relax European Association of Geoscientists \& Engineers, 2016, pp. 1--5.

\bibitem{cruz2021tupi}
N.~M. Cruz, J.~M. Cruz, L.~M. Teixeira, M.~M. da~Costa, L.~B. de~Oliveira, E.~N. Urasaki, T.~P. Bispo, M.~de~S{\'a}~Jardim, M.~H. Grochau, and A.~Maul, ``Tupi nodes pilot: A successful 4d seismic case for brazilian presalt reservoirs,'' \emph{The Leading Edge}, vol.~40, no.~12, pp. 886--896, 2021.

\bibitem{fink2000time}
M.~Fink, D.~Cassereau, A.~Derode, C.~Prada, P.~Roux, M.~Tanter, J.-L. Thomas, and F.~Wu, ``Time-reversed acoustics,'' \emph{Reports on progress in Physics}, vol.~63, no.~12, p. 1933, 2000.

\bibitem{Ultrasound-2024-Li}
T.~Li, B.~Feng, Y.~Zhou, R.~Xu, and H.~Wang, \emph{Ultrasound reverse time migration for human brain imaging}, 2024, pp. 2269--2273.

\bibitem{Truncated-2024-Wu}
F.~Wu, Q.~He, Y.~Li, B.~Han, and Y.~Wang, ``Truncated newton full waveform inversion method for the human brain imaging,'' \emph{Journal of Physics: Conference Series}, vol. 2822, no.~1, p. 012013, sep 2024.

\bibitem{Amortized-2023-Orozco}
R.~Orozco, M.~Louboutin, A.~Siahkoohi, G.~Rizzuti, T.~van Leeuwen, and F.~Herrmann, ``Amortized normalizing flows for transcranial ultrasound with uncertainty quantification,'' in \emph{Proceedings of Machine Learning Research}, 2023, pp. 332--349.

\bibitem{Learned-2023-Lozenski}
L.~Lozenski, H.~Wang, F.~Li, M.~Anastasio, B.~Wohlberg, Y.~Lin, and U.~Villa, ``Learned full waveform inversion incorporating task information for ultrasound computed tomography,'' \emph{IEEE Transactions on Computational Imaging}, vol.~10, pp. 69--82, 2024.

\bibitem{fan2022model}
Y.~Fan, H.~Wang, H.~Gemmeke, T.~Hopp, and J.~Hesser, ``Model-data-driven image reconstruction with neural networks for ultrasound computed tomography breast imaging,'' \emph{Neurocomputing}, vol. 467, pp. 10--21, 2022.

\bibitem{roy2016ultrasound}
O.~Roy, M.~Zuberi, R.~Pratt, and N.~Duric, ``Ultrasound breast imaging using frequency domain reverse time migration,'' in \emph{Medical Imaging 2016: Ultrasonic Imaging and Tomography}, vol. 9790.\hskip 1em plus 0.5em minus 0.4em\relax SPIE, 2016, pp. 84--92.

\bibitem{Waveform-2015-Wang}
K.~Wang, T.~Matthews, F.~Anis, C.~Li, N.~Duric, and M.~Anastasio, ``Waveform inversion with source encoding for breast sound speed reconstruction in ultrasound computed tomography,'' \emph{IEEE Trans Ultrason Ferroelectr Freq Control}, vol.~62, no.~3, pp. 475--493, 2015.

\bibitem{agudo20183d}
O.~C. Agudo, L.~Guasch, P.~Huthwaite, and M.~Warner, ``3d imaging of the breast using full-waveform inversion,'' in \emph{Proc. Int. Workshop Med. Ultrasound Tomogr}, 2018, pp. 99--110.

\bibitem{Openpros-2025-Wang}
\BIBentryALTinterwordspacing
H.~Wang, Y.~Wu, Y.~Feng, P.~Jin, S.~Feng, L.~Zhang, J.~Wiskin, B.~Turkbey, P.~A. Pinto, B.~J. Wood, S.~Luo, Y.~Chen, E.~Boctor, and Y.~Lin, ``Openpros: A large-scale dataset for limited view prostate ultrasound computed tomography,'' 2025. [Online]. Available: \url{https://arxiv.org/abs/2505.12261}
\BIBentrySTDinterwordspacing

\bibitem{New-2024-Pan}
Y.~Pan, X.~Wang, Y.~Qiang, N.~Wang, R.~Liu, G.~Yang, Z.~Zhang, X.~He, Y.~Yu, H.~Zheng, and W.~Qiu, ``A new method of plane-wave ultrasound imaging based on reverse time migration,'' \emph{IEEE Trans Biomed Eng.}, vol.~71, no.~5, pp. 1628--1639, 2024.

\bibitem{wang2015reverse}
Z.~Wang, H.~Ding, G.~Lu, and X.~Bi, ``Reverse-time migration based optical imaging,'' \emph{IEEE Transactions on medical imaging}, vol.~35, no.~1, pp. 273--281, 2015.

\bibitem{parmar2018image}
N.~Parmar, A.~Vaswani, J.~Uszkoreit, L.~Kaiser, N.~Shazeer, A.~Ku, and D.~Tran, ``Image transformer,'' in \emph{International conference on machine learning}.\hskip 1em plus 0.5em minus 0.4em\relax PMLR, 2018, pp. 4055--4064.

\bibitem{han2022survey}
K.~Han, Y.~Wang, H.~Chen, X.~Chen, J.~Guo, Z.~Liu, Y.~Tang, A.~Xiao, C.~Xu, Y.~Xu \emph{et~al.}, ``A survey on vision transformer,'' \emph{IEEE transactions on pattern analysis and machine intelligence}, vol.~45, no.~1, pp. 87--110, 2022.

\bibitem{lu2022transformer}
Z.~Lu, J.~Li, H.~Liu, C.~Huang, L.~Zhang, and T.~Zeng, ``Transformer for single image super-resolution,'' in \emph{Proceedings of the IEEE/CVF conference on computer vision and pattern recognition}, 2022, pp. 457--466.

\bibitem{yang2020learning}
F.~Yang, H.~Yang, J.~Fu, H.~Lu, and B.~Guo, ``Learning texture transformer network for image super-resolution,'' in \emph{Proceedings of the IEEE/CVF conference on computer vision and pattern recognition}, 2020, pp. 5791--5800.

\bibitem{li2022hst}
B.~Li, X.~Li, Y.~Lu, S.~Liu, R.~Feng, and Z.~Chen, ``Hst: Hierarchical swin transformer for compressed image super-resolution,'' in \emph{European Conference on Computer Vision}.\hskip 1em plus 0.5em minus 0.4em\relax Springer, 2022, pp. 651--668.

\bibitem{kim2022instaformer}
S.~Kim, J.~Baek, J.~Park, G.~Kim, and S.~Kim, ``Instaformer: Instance-aware image-to-image translation with transformer,'' in \emph{Proceedings of the IEEE/CVF Conference on Computer Vision and Pattern Recognition}, 2022, pp. 18\,321--18\,331.

\bibitem{torbunov2023uvcgan}
D.~Torbunov, Y.~Huang, H.~Yu, J.~Huang, S.~Yoo, M.~Lin, B.~Viren, and Y.~Ren, ``Uvcgan: Unet vision transformer cycle-consistent gan for unpaired image-to-image translation,'' in \emph{Proceedings of the IEEE/CVF Winter Conference on Applications of Computer Vision}, 2023, pp. 702--712.

\bibitem{zheng2022ittr}
W.~Zheng, Q.~Li, G.~Zhang, P.~Wan, and Z.~Wang, ``Ittr: Unpaired image-to-image translation with transformers,'' \emph{arXiv preprint arXiv:2203.16015}, 2022.

\bibitem{gao2021utnet}
Y.~Gao, M.~Zhou, and D.~N. Metaxas, ``Utnet: a hybrid transformer architecture for medical image segmentation,'' in \emph{Medical Image Computing and Computer Assisted Intervention--MICCAI 2021: 24th International Conference, Strasbourg, France, September 27--October 1, 2021, Proceedings, Part III 24}.\hskip 1em plus 0.5em minus 0.4em\relax Springer, 2021, pp. 61--71.

\bibitem{huang2021missformer}
X.~Huang, Z.~Deng, D.~Li, and X.~Yuan, ``Missformer: An effective medical image segmentation transformer,'' \emph{arXiv preprint arXiv:2109.07162}, 2021.

\bibitem{jain2023oneformer}
J.~Jain, J.~Li, M.~T. Chiu, A.~Hassani, N.~Orlov, and H.~Shi, ``Oneformer: One transformer to rule universal image segmentation,'' in \emph{Proceedings of the IEEE/CVF Conference on Computer Vision and Pattern Recognition}, 2023, pp. 2989--2998.

\bibitem{shamshad2023transformers}
F.~Shamshad, S.~Khan, S.~W. Zamir, M.~H. Khan, M.~Hayat, F.~S. Khan, and H.~Fu, ``Transformers in medical imaging: A survey,'' \emph{Medical Image Analysis}, p. 102802, 2023.

\bibitem{albawi2017understanding}
S.~Albawi, T.~A. Mohammed, and S.~Al-Zawi, ``Understanding of a convolutional neural network,'' in \emph{2017 international conference on engineering and technology (ICET)}.\hskip 1em plus 0.5em minus 0.4em\relax IEEE, 2017, pp. 1--6.

\bibitem{dong2014learning}
C.~Dong, C.~C. Loy, K.~He, and X.~Tang, ``Learning a deep convolutional network for image super-resolution,'' in \emph{Computer Vision--ECCV 2014: 13th European Conference, Zurich, Switzerland, September 6-12, 2014, Proceedings, Part IV 13}.\hskip 1em plus 0.5em minus 0.4em\relax Springer, 2014, pp. 184--199.

\bibitem{Duong2021}
V.~Van~Duong, T.~N. Huu, J.~Yim, and B.~Jeon, ``A fast and efficient super-resolution network using hierarchical dense residual learning,'' in \emph{2021 IEEE International Conference on Image Processing (ICIP)}.\hskip 1em plus 0.5em minus 0.4em\relax IEEE, 2021, pp. 1809--1813.

\bibitem{Dai2019}
T.~Dai, J.~Cai, Y.~Zhang, S.-T. Xia, and L.~Zhang, ``Second-order attention network for single image super-resolution,'' in \emph{2019 IEEE/CVF Conference on Computer Vision and Pattern Recognition (CVPR)}, 2019, pp. 11\,057--11\,066.

\bibitem{zhang2018residual}
Y.~Zhang, Y.~Tian, Y.~Kong, B.~Zhong, and Y.~Fu, ``Residual dense network for image super-resolution,'' \emph{arXiv preprint arXiv:1802.08797}, 2018.

\bibitem{ahn2018fast}
N.~Ahn, B.~Kang, and K.-A. Sohn, ``Fast, accurate, and lightweight super-resolution with cascading residual network,'' in \emph{Proceedings of the European conference on computer vision (ECCV)}, 2018, pp. 252--268.

\bibitem{Tai2017}
Y.~Tai, J.~Yang, and X.~Liu, ``Image super-resolution via deep recursive residual network,'' in \emph{Proceedings of the IEEE conference on computer vision and pattern recognition}, 2017, pp. 3147--3155.

\bibitem{ledig2017photo}
C.~Ledig, L.~Theis, F.~Husz{\'a}r, J.~Caballero, A.~Cunningham, A.~Acosta, A.~Aitken, A.~Tejani, J.~Totz, Z.~Wang \emph{et~al.}, ``Photo-realistic single image super-resolution using a generative adversarial network,'' in \emph{Proceedings of the IEEE conference on computer vision and pattern recognition}, 2017, pp. 4681--4690.

\bibitem{chen2017fsrnet}
Y.~Chen, Y.~Tai, X.~Liu, C.~Shen, and J.~Yang, ``Fsrnet: End-to-end learning face super-resolution with facial priors,'' in \emph{Proceedings of the IEEE conference on computer vision and pattern recognition}, 2018, pp. 2492--2501.

\bibitem{wang2018recovering}
X.~Wang, K.~Yu, C.~Dong, and C.~C. Loy, ``Recovering realistic texture in image super-resolution by deep spatial feature transform,'' in \emph{Proceedings of the IEEE conference on computer vision and pattern recognition}, 2018, pp. 606--615.

\bibitem{wang2018esrgan}
X.~Wang, K.~Yu, S.~Wu, J.~Gu, Y.~Liu, C.~Dong, Y.~Qiao, and C.~Change~Loy, ``Esrgan: Enhanced super-resolution generative adversarial networks,'' in \emph{ECCV Workshops}, 2018.

\bibitem{karras2018progressive}
T.~Karras, T.~Aila, S.~Laine, and J.~Lehtinen, ``Progressive growing of gans for improved quality, stability, and variation,'' \emph{arXiv preprint arXiv:1710.10196}, 2017.

\bibitem{gan8759375}
U.~Upadhyay and S.~P. Awate, ``Robust super-resolution gan, with manifold-based and perception loss,'' in \emph{2019 IEEE 16th International Symposium on Biomedical Imaging (ISBI 2019)}.\hskip 1em plus 0.5em minus 0.4em\relax IEEE, 2019, pp. 1372--1376.

\bibitem{cheng2021mfagan}
W.~Cheng, M.~Zhao, Z.~Ye, and S.~Gu, ``Mfagan: A compression framework for memory-efficient on-device super-resolution gan,'' \emph{arXiv preprint arXiv:2107.12679}, 2021.

\bibitem{Long2021}
Z.~Wenlong, L.~Yihao, C.~Dong, and Y.~Qiao, ``Ranksrgan: Generative adversarial networks with ranker for image super-resolution,'' \emph{IEEE Transactions on Pattern Analysis and Machine Intelligence}, vol.~44, no.~10, pp. 1--1, 2021.

\bibitem{fang2022cross}
C.~Fang, D.~Zhang, L.~Wang, Y.~Zhang, L.~Cheng, and J.~Han, ``Cross-modality high-frequency transformer for mr image super-resolution,'' in \emph{Proceedings of the 30th ACM International Conference on Multimedia}, 2022, pp. 1584--1592.

\bibitem{fang2022hybrid}
J.~Fang, H.~Lin, X.~Chen, and K.~Zeng, ``A hybrid network of cnn and transformer for lightweight image super-resolution,'' in \emph{Proceedings of the IEEE/CVF conference on computer vision and pattern recognition}, 2022, pp. 1103--1112.

\bibitem{wang2022detail}
S.~Wang, T.~Zhou, Y.~Lu, and H.~Di, ``Detail-preserving transformer for light field image super-resolution,'' in \emph{Proceedings of the AAAI Conference on Artificial Intelligence}, vol.~36, no.~3, 2022, pp. 2522--2530.

\bibitem{gao2023implicit}
S.~Gao, X.~Liu, B.~Zeng, S.~Xu, Y.~Li, X.~Luo, J.~Liu, X.~Zhen, and B.~Zhang, ``Implicit diffusion models for continuous super-resolution,'' in \emph{Proceedings of the IEEE/CVF conference on computer vision and pattern recognition}, 2023, pp. 10\,021--10\,030.

\bibitem{li2022srdiff}
H.~Li, Y.~Yang, M.~Chang, S.~Chen, H.~Feng, Z.~Xu, Q.~Li, and Y.~Chen, ``Srdiff: Single image super-resolution with diffusion probabilistic models,'' \emph{Neurocomputing}, vol. 479, pp. 47--59, 2022.

\bibitem{moser2024diffusion}
B.~B. Moser, A.~S. Shanbhag, F.~Raue, S.~Frolov, S.~Palacio, and A.~Dengel, ``Diffusion models, image super-resolution, and everything: A survey,'' \emph{IEEE Transactions on Neural Networks and Learning Systems}, 2024.

\bibitem{lozenski2024learned}
L.~Lozenski, H.~Wang, F.~Li, M.~Anastasio, B.~Wohlberg, Y.~Lin, and U.~Villa, ``Learned full waveform inversion incorporating task information for ultrasound computed tomography,'' \emph{IEEE transactions on computational imaging}, vol.~10, pp. 69--82, 2024.

\bibitem{zhang2024cross}
N.~Zhang, Y.~Zhao, Y.~Yuan, Y.~Xiao, M.~Qin, and Y.~Shen, ``Cross-correlation adjustment full-waveform inversion with source encoding in ultrasound computed tomography,'' \emph{Ultrasonics}, vol. 142, p. 107392, 2024.

\bibitem{vargasbreast}
A.~Vargas, N.~Hernandez, A.~B. Ramirez, and S.~Pertuz, ``Breast cancer detection from ultrasound computed tomography imaging using radiomic analysis: in silico trial,'' \emph{Medical Physics}.

\bibitem{sharon2024real}
T.~Sharon and Y.~C. Eldar, ``Real-time model-based quantitative ultrasound and radar,'' \emph{IEEE Transactions on Computational Imaging}, 2024.

\bibitem{goyal2025ultrasound}
R.~Goyal, T.~Rymarczyk, and M.~Soleimani, ``Ultrasound computed tomography for in-vivo lung imaging.'' \emph{IEEE Transactions on Instrumentation and Measurement}, 2025.

\bibitem{marty2021acoustoelastic}
P.~Marty, C.~Boehm, and A.~Fichtner, ``Acoustoelastic full-waveform inversion for transcranial ultrasound computed tomography,'' in \emph{Medical Imaging 2021: Ultrasonic Imaging and Tomography}, vol. 11602.\hskip 1em plus 0.5em minus 0.4em\relax SPIE, 2021, pp. 210--229.

\bibitem{fink1992time}
M.~Fink, ``Time reversal of ultrasonic fields. i. basic principles,'' \emph{IEEE transactions on ultrasonics, ferroelectrics, and frequency control}, vol.~39, no.~5, pp. 555--566, 1992.

\bibitem{zhang2021deep}
W.~Zhang and J.~Gao, ``Deep-learning full-waveform inversion using seismic migration images,'' \emph{IEEE Transactions on Geoscience and Remote Sensing}, vol.~60, pp. 1--18, 2021.

\bibitem{zhang2021least}
W.~Zhang, J.~Gao, T.~Yang, X.~Jiang, and W.~Sun, ``Least-squares reverse time migration using convolutional neural networks,'' \emph{Geophysics}, vol.~86, no.~6, pp. R959--R971, 2021.

\bibitem{geng2022deep}
Z.~Geng, Z.~Zhao, Y.~Shi, X.~Wu, S.~Fomel, and M.~Sen, ``Deep learning for velocity model building with common-image gather volumes,'' \emph{Geophysical Journal International}, vol. 228, no.~2, pp. 1054--1070, 2022.

\bibitem{xu2024utilizing}
W.~Xu and X.~Wul, ``Utilizing message passing mechanism,'' in \emph{Big Data: 12th CCF Conference, BigData 2024, Qingdao, China, August 9--11, 2024, Proceedings}.\hskip 1em plus 0.5em minus 0.4em\relax Springer Nature, p.~62.

\bibitem{barkau1996unet}
R.~L. Barkau, ``Unet: One-dimensional unsteady flow through a full network of open channels. user's manual,'' Hydrologic Engineering Center Davis CA, Tech. Rep., 1996.

\bibitem{zhang2018image_backup}
Y.~Zhang, K.~Li, K.~Li, L.~Wang, B.~Zhong, and Y.~Fu, ``Image super-resolution using very deep residual channel attention networks,'' in \emph{ECCV}, 2018.

\bibitem{chu2017cyclegan}
C.~Chu, A.~Zhmoginov, and M.~Sandler, ``Cyclegan, a master of steganography,'' \emph{arXiv preprint arXiv:1712.02950}, 2017.

\bibitem{wang2004image}
Z.~Wang, A.~C. Bovik, H.~R. Sheikh, and E.~P. Simoncelli, ``Image quality assessment: from error visibility to structural similarity,'' \emph{IEEE transactions on image processing}, vol.~13, no.~4, pp. 600--612, 2004.

\bibitem{kingma2014adam_arxiv}
D.~P. Kingma and J.~Ba, ``Adam: A method for stochastic optimization,'' \emph{arXiv preprint arXiv:1412.6980}, 2014.

\end{thebibliography}

\end{document}